\documentclass[final,5p,times,twocolumn]{elsarticle}

\usepackage{algorithm}
\usepackage{algorithmic}
\usepackage{wrapfig}
\usepackage{subfigure}
\usepackage{multirow}
\usepackage{amssymb}
\usepackage{amsmath}
\usepackage[table,xcdraw]{xcolor}
\usepackage{natbib}
\setcitestyle{sort&compress}
\usepackage[colorlinks,
            linkcolor=red,
            anchorcolor=blue,
            citecolor=green
            ]{hyperref}
\usepackage{graphicx}
\usepackage{booktabs}
\usepackage{makecell}
\usepackage{url}
\usepackage{array}
\usepackage{pifont}

\newcommand{\mygray}[1]{\textcolor{gray}{#1}}
\newcommand{\revised}[1]{\textcolor{black}{#1}}

\journal{Knowledge-Based Systems}

\begin{document}

\begin{frontmatter}

%% Title, authors and addresses

%% use the tnoteref command within \title for footnotes;
%% use the tnotetext command for theassociated footnote;
%% use the fnref command within \author or \affiliation for footnotes;
%% use the fntext command for theassociated footnote;
%% use the corref command within \author for corresponding author footnotes;
%% use the cortext command for theassociated footnote;
%% use the ead command for the email address,
%% and the form \ead[url] for the home page:
%% \title{Title\tnoteref{label1}}
%% \tnotetext[label1]{}
%% \author{Name\corref{cor1}\fnref{label2}}
%% \ead{email address}
%% \ead[url]{home page}
%% \fntext[label2]{}
%% \cortext[cor1]{}
%% \affiliation{organization={},
%%            addressline={}, 
%%            city={},
%%            postcode={}, 
%%            state={},
%%            country={}}
%% \fntext[label3]{}

\title{Expression Prompt Collaboration Transformer for Universal Referring\\Video Object Segmentation}

%% use optional labels to link authors explicitly to addresses:
%% \author[label1,label2]{}
%% \affiliation[label1]{organization={},
%%             addressline={},
%%             city={},
%%             postcode={},
%%             state={},
%%             country={}}
%%
%% \affiliation[label2]{organization={},
%%             addressline={},
%%             city={},
%%             postcode={},
%%             state={},
%%             country={}}

\author[1]{Jiajun Chen}
\ead{chenjiajun@hnu.edu.cn}
 
\author[2]{Jiacheng Lin}
\ead{jcheng_lin@hnu.edu.cn}

\author[2]{Guojin Zhong}
\ead{gjzhong@hnu.edu.cn}

\author[2]{Haolong Fu}
\ead{haolongfu@hnu.edu.cn}

\author[3]{Ke Nai}
\ead{naike_hnu@hnu.edu.cn}

\author[1]{Kailun Yang\corref{cor1}}
\ead{kailun.yang@hnu.edu.cn}

\author[1,2]{Zhiyong Li\corref{cor1}}
\ead{zhiyong.li@hnu.edu.cn}

\address[1]{School of Robotics, Hunan University, Changsha 410082, China.}
 
\address[2]{College of Computer Science and Electronic Engineering, Hunan University, Changsha 410082, China.}

\address[3]{School of Computer and Communication Engineering, Changsha University of Science and Technology, Changsha 410011, China.}

\cortext[cor1]{Corresponding author at: School of Robotics, Hunan University, Changsha 410082, China}

\begin{abstract}
%% Text of abstract

Audio-guided Video Object Segmentation (A-VOS) and Referring Video Object Segmentation (R-VOS) are two highly related tasks that both aim to segment specific objects from video sequences according to expression prompts. However, due to the challenges of modeling representations for different modalities, existing methods struggle to strike a balance between interaction flexibility and localization precision. In this paper, we address this problem from two perspectives: the alignment of audio and text and the deep interaction among audio, text, and visual modalities. First, we propose a universal architecture, the Expression Prompt Collaboration Transformer, herein EPCFormer. Next, we propose an Expression Alignment (EA) mechanism for audio and text. The proposed EPCFormer exploits the fact that audio and text prompts referring to the same objects are semantically equivalent by using contrastive learning for both types of expressions. Then, to facilitate deep interactions among audio, text, and visual modalities, we introduce an Expression-Visual Attention (EVA) module. The knowledge of video object segmentation in terms of the expression prompts can seamlessly transfer between the two tasks by deeply exploring complementary cues between text and audio. Experiments on well-recognized benchmarks demonstrate that our  EPCFormer attains state-of-the-art results on both tasks. The source code will be made publicly available at \url{https://github.com/lab206/EPCFormer}.
\end{abstract}

\begin{keyword}
% keywords here, in the form: keyword \sep keyword
Audio-guide video object segmentation \sep Referring video object segmentation \sep Expression-visual attention \sep Audio-text contrastive learning \sep Multi-task learning.
\end{keyword}

\end{frontmatter}

%% \linenumbers

%% main text
\section{Introduction}
\label{sec:introduction}

\revised{Audio-gudied Video Object Segmentation (A-VOS)~\cite{pan2022wnet} and Referring Video Object Segmentation (R-VOS)}~\cite{GavrilyukGLS18,MTTR,ReferFormer} aim to segment specific objects from video sequences by a \revised{given audio or text prompt.} \revised{They can be used in many application situations,} \revised{\textit{e.g.},} video editing~\cite{zhang2024video} \revised{and human-computer interaction~\cite{nake2001human,xie2024satr}.} Currently, both tasks have been widely discussed in various research fields, \revised{\textit{e.g.},} expression-video fusion~\cite{pan2022wnet,wang2019acan}, encoder-decoder design~\cite{GavrilyukGLS18,MTTR}, and referring localization~\cite{khoreva2018video,jing2021locate}, leading to significant advancements.

\begin{figure}[tb!]
\centering
\includegraphics[width=0.45\textwidth]{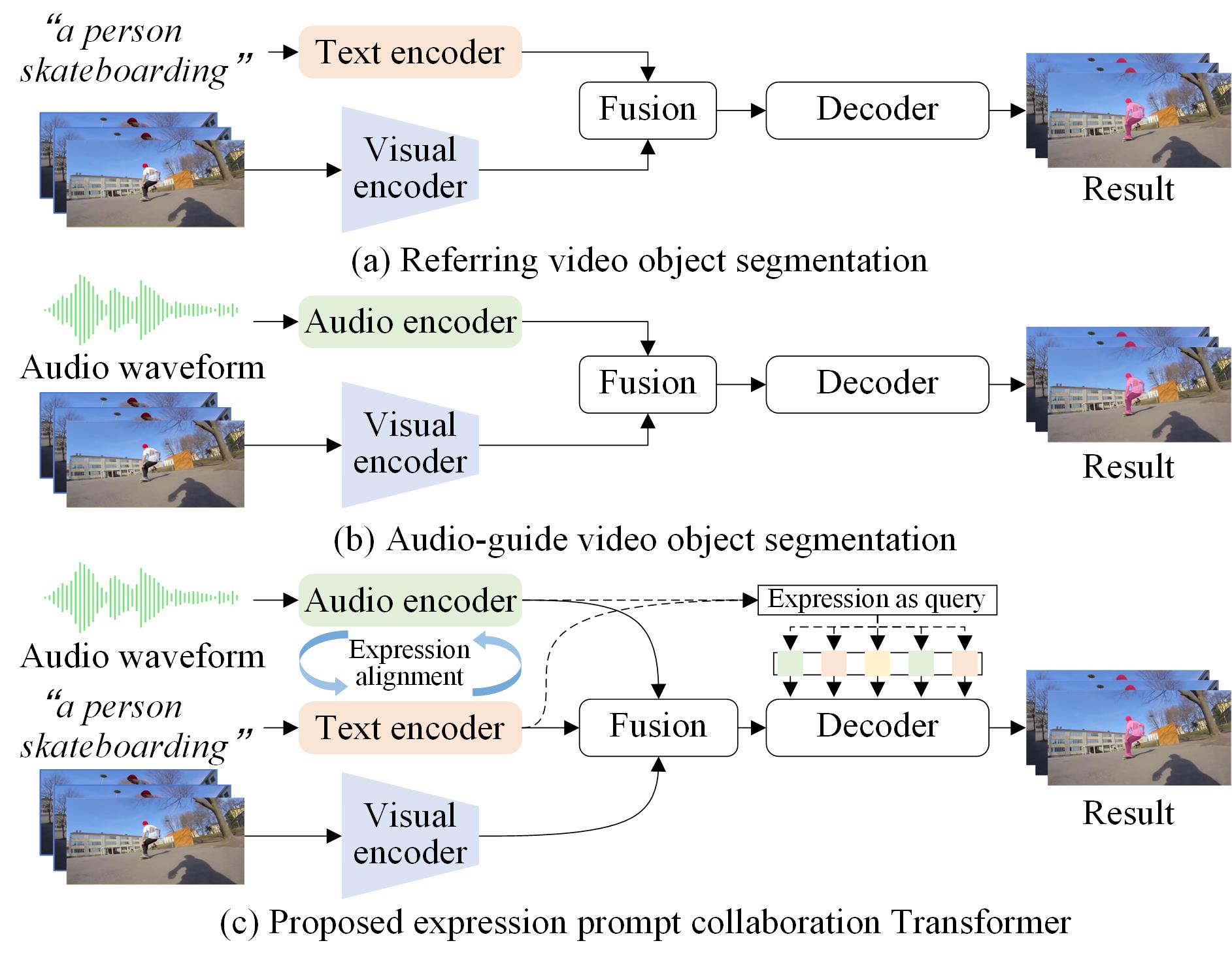}
\caption{\revised{Comparison of EPCFormer and existing models.
(a) A typical text-guided model for R-VOS.
(b) A typical audio-guided model for A-VOS.
(c) Our EPCFormer, which is capable of processing both text and audio prompts.}}
\label{fig:intro}
\end{figure}

R-VOS is shown in \revised{Figure}~\ref{fig:intro}(a), which has been widely studied due to its high-precision localization capabilities.
\revised{However}, recent work~\cite{pan2022wnet} suggests that R-VOS may not be sufficiently efficient for practical applications.
Notoriously, obtaining text clues in many real-world scenarios is difficult, whereas using audio aligns better with human-computer interaction.
In light of these factors, recent works~\cite{pan2022wnet, LIN2023120960} attempt to use Automatic Speech Recognition (ASR)~\cite{schneider2019wav2vec,hsu2021hubert} to transcribe audio prompts for R-VOS. 
\revised{Despite their progress, it is challenging for these methods to achieve optimal results due to inevitable translation errors and redundant computation costs.}
To address these issues, Pan~\textit{et al.}~\cite{pan2022wnet} propose an A-VOS paradigm, as shown in \revised{Figure}~\ref{fig:intro}(b), which directly fuses the audio and visual features to segment the referred object in the video. 
Although it has high interaction flexibility, it still faces challenges, such as the difficulty of audio-visual feature alignment and fusion~\cite{pan2022wnet,ning2022audio}, leading to imprecise results in referred object localization~\cite{zhou2022audio,gao2023avsegformer}.

To address these challenges, we introduce a novel universal architecture, Expression Prompt Collaboration Transformer (EPCFormer), \revised{to learn and process text and audio prompts simultaneously.} 
On the one hand, the features from one modality can be refined based on the knowledge learned from another modality, and vice versa, enhancing the integration and comprehension of the multi-modal data.
To promote learning of the model and narrow the gap when processing these two modalities, we bridge audio and text domains via an efficient supervision mechanism based on contrastive learning, termed Expression Alignment (EA). 
\revised{The audio and text features are projected into a multi-modal embedding space through an EA mechanism, where referring semantics are aggregated.}
\revised{In this way, the model better exploits the semantic equivalence between audio and text prompts depicting the same objects.}

\revised{On the other hand, we proposed an Expression-Visual Attention (EVA) module with audio-text collaboration block and expression-visual interaction block. This module is designed to implement composable interactions among three types of modalities within a unified network.} It \revised{enables} the model \revised{to handle audio-only, text-only, or combined audio-text prompts in a unified manner.}
\revised{The proposed model has two parallel process pathways, as shown in Figure~\ref{fig:intro}(c).} 
One pathway processes audio prompts, while the other one processes text prompts.
Under the designed multi-task training method, two types of referring features are densely integrated.
\revised{As a result}, the model is encouraged to learn a united multi-modal representation for visual and two types of referring features.
In this way, the model effectively emphasizes matching features of visual regions and crucial elements of the referring expressions\revised{, while also} establishing complementary connections between audio and text features.
Experimental results on seven benchmarks demonstrate that the proposed EPCFormer achieves better or comparable results against state-of-the-art methods.

At a glance, this work delivers the following contributions:

1) \revised{We propose an Expression Prompt Collaboration Transformer (EPCFormer) for R-VOS and A-VOS tasks.} EPCFormer leverages audio and text as prompts to effectively segment the referred objects in the video, achieving high-precision localization and exceptional interaction flexibility.

2) We propose an Expression Alignment (EA) mechanism that enables effective semantic-level contrastive learning between audio and text features and narrows the gap when processing these two modalities.

3) \revised{We propose an Expression-Visual Attention (EVA) module to handle interactions between audio or text cues and video independently or jointly and make connections between audio and text features that work effectively together.}

The subsequent sections of this paper are organized as follows: \revised{Section}~\ref{sec:related-work} provides a brief overview of related work.
The proposed methods are described in detail in \revised{Section}~\ref{sec:method}.
Experimental results are presented and discussed in \revised{Section}~\ref{sec:experiments}, followed by the conclusion in \revised{Section}~\ref{sec:conclusion}.

\section{Related Work} \label{sec:related-work}

\subsection{Referring Video Object Segmentation}
\label{subsec:rvos}

\textbf{Text-guided video object segmentation.} R-VOS refers to segmenting specific objects from video frames based on the given text prompts~\cite{
URVOS,liu2022cmpc,miao2023spectrum,hui2021collaborative,liang2021clawcranenet,YOFO,wu2022multi}.
Gavrilyuk~\textit{et al.}~\cite{GavrilyukGLS18} first explore R-VOS and propose to encode linguistic clues as dynamic filters for visual features. 
To handle complex sentences, subsequent works widely adopt cross-modal attention mechanisms~\cite{vaswani2017attention}. 
\revised{For example, Wang~\textit{et al.}~\cite{wang2019acan} employ an asymmetric cross-modal attention mechanism.} 
Ning~\textit{et al.}~\cite{prpe} introduce polar positional encoding and polar attention module to enhance the representation of positional relations in the text.
\revised{To explore incorporating temporal cues and boost performance, Ye~\textit{et al.}~\cite{ye2021referring} propose a cross-frame self-attention module to capture the temporal context in consecutive frames.} 
Ding~\textit{et al.}~\cite{LBDT} adopt a dual-stream architecture to highlight the spatial-temporal features.

Recently, Transformer-based methods have been used in R-VOS. 
\revised{For instance}, Ding~\textit{et al.}~\cite{vlt} employ referring text to generate dynamic queries. 
\revised{MTTR~\cite{MTTR} employs an instance-level segmentation transformer inspired by~\cite{DETR,DeformableDETR,VISTR}.} ReferFormer~\cite{ReferFormer} leverages the linguistic prompts as decoder queries to attend to relevant regions in video frames. Most recently, Wu~\textit{et al.}~\cite{wu2023online} \revised{designed} a cross-frame query propagation to transform matching instance queries into subsequent frames.

\textbf{Audio-guided video object segmentation.}
A-VOS aims to predict a sequence of segmentation masks according to given audio prompts. 
Pan~\textit{et al.}~\cite{pan2022wnet} pioneer this task and leverage a Transformer model with an audio-visual cross-modal attention module to capture the intricate semantic representations of audio-video interactions.
In addition to A-VOS, \revised{recent works predominantly center around Audio-Visual Segmentation (AVS)~\cite{zhou2022audio}, and segments sounding objects corresponding to the given sound.}
\revised{Zhou~\textit{et al.}~\cite{zhou2022audio} utilize cross-modal attention to exchange information between visual and acoustic features.}
Gao~\textit{et al.}~\cite{gao2023avsegformer} employ the audio as queries for Transformers~\cite{DETR, Mask2Former}~to focus on distinctive features of sounding objects.
More recently, some research~\cite{yan2023referred} has explored the unified model of R-VOS and AVS tasks and achieved impressive results. However, it cannot promote mutual learning between text and audio, so it cannot be directly used in this task. Despite the pioneering success, existing methods are not efficient enough to model the semantic representations of audio, text, and visual interaction contents.

Unlike existing methods, the proposed method can handle both text and audio for a broader range of applications.
\revised{Additionally, expression-visual attention ensures that three different modalities, \textit{i.e.}, audio, text, and visual, achieve effective interactions and complementary information exchange between two different referring prompts.}

\subsection{Contrastive Learning for Multi-modal Alignment}
\label{subsec:contractive_learning}

Contrastive learning~\cite{hadsell2006dimensionality,CLIP}, a pivotal aspect of deep learning~\cite{ying2023ctvis,zhong2023contrast}, 
initially models image similarity and dissimilarity across two or more perspectives~\cite{hadsell2006dimensionality}.
Recently, the methodology has been expanded to include video segmentation~\cite{ying2023ctvis,IDOL}, referring segmentation~\cite{vlt,cris,luo2023soc}, \revised{and} audio-text alignment~\cite{CLAP,bapna2022mslam,zhang2022speechlm}. 
For example, CTVIS~\cite{ying2023ctvis} utilizes contrastive loss to associate discriminative instance-level features in the multi-frame.
Luo~\textit{et al.}~\cite{luo2023soc} cluster video-level object representations with linguistic features via contrastive loss.
Spurred by the success of aligning visual and language features via contrastive learning~\cite{CLIP,jia2021scaling},
CLAP~\cite{CLAP} and related work, \textit{e.g.},~\cite{bapna2022mslam,zhang2022speechlm}, popularize learning audio representations through language supervision.
They demonstrate that effective audio-text alignment enables the pipeline to achieve impressive results in text-audio downstream tasks.
Furthermore, to enforce the tri-modal alignment~\cite{zhu2023vatlm,guzhov2022audioclip}, Shih~\textit{et al.}~\cite{SpeechCLIP} realize bridging audio and text domains via image backbone~\cite{CLIP} without transcriptions, 
whereas Guzhov~\textit{et al.}~\cite{guzhov2022audioclip} simultaneously learn a joint representation for image, text, and audio modalities.

%In this work, we endeavor to investigate unified representational referring prompts for R-VOS and A-VOS.  This exploration is centered on ensuring that the semantics of text and audio can be seamlessly aligned within the feature space.

\revised{Building upon the advances of contrastive learning, we introduce an expression alignment mechanism. This mechanism enables the model to align features of different modality prompts related to the same object, maximizing their similarity within the representation space.}

\subsection{Universal Visual Segmentation}
\label{subsec:universal}

\revised{The goal of universal visual segmentation is to combine multiple segmentation tasks into a single model~\cite{zhang2023cmx,zhang2023delivering,gu2023dataseg,zhang2023vpuformer,wang2023seggpt,wu2023segment_every_reference,GLIP}.}
%Through a shared backbone, universal knowledge transfer between tasks becomes effortlessly attainable~\cite{kolesnikov2020big,chen2022unified}.
\revised{For instance, K-Net~\cite{zhang2021k} uses a group of dynamic learnable kernels to unify instance, semantic, and panoptic segmentation, whereas Mask2Former~\cite{Mask2Former} builds on~\cite{cheng2021maskformer} and improves it at different segmentation tasks by adding learnable queries and a masked cross-attention mechanism.}
MCN~\cite{MCN} achieves collaborative learning of referring expression comprehension and segmentation.
OneFormer~\cite{jain2023oneformer} handles different segmentation predictions with a task-conditioned joint training strategy using a unified set of object queries for guidance.
MaskDINO~\cite{MaskDINO} aims to unify segmentation and detection, whereas HIPIE~\cite{wang2023hipie} jointly learns an open-vocabulary model for both tasks.
Recently, some works~\cite{UNINEXT,wu2023glee} innovatively convert task-specific preconditions into prompts. 
For instance, UNINEXT~\cite{UNINEXT} employs a prompt generation paradigm to address ten instance perception tasks. 
UniLSeg~\cite{liu2023universal} performs scene segmentation at arbitrary granularity or semantic level using language instructions as guidance.
Additionally, SAM~\cite{sam} and SEEM~\cite{SEEM} engage in image segmentation with diverse user interactions facilitated through prompts.

\revised{Unfortunately, existing works lack an effective representation to integrate these two tasks. In this work, we present a universal architecture to handle R-VOS and A-VOS simultaneously. This architecture learns to maintain a consistent understanding of audio and text prompts, seamlessly transferring the learned universal knowledge to both tasks.}

\begin{figure*}[t!]
\centering
\includegraphics[width=1.0\linewidth]{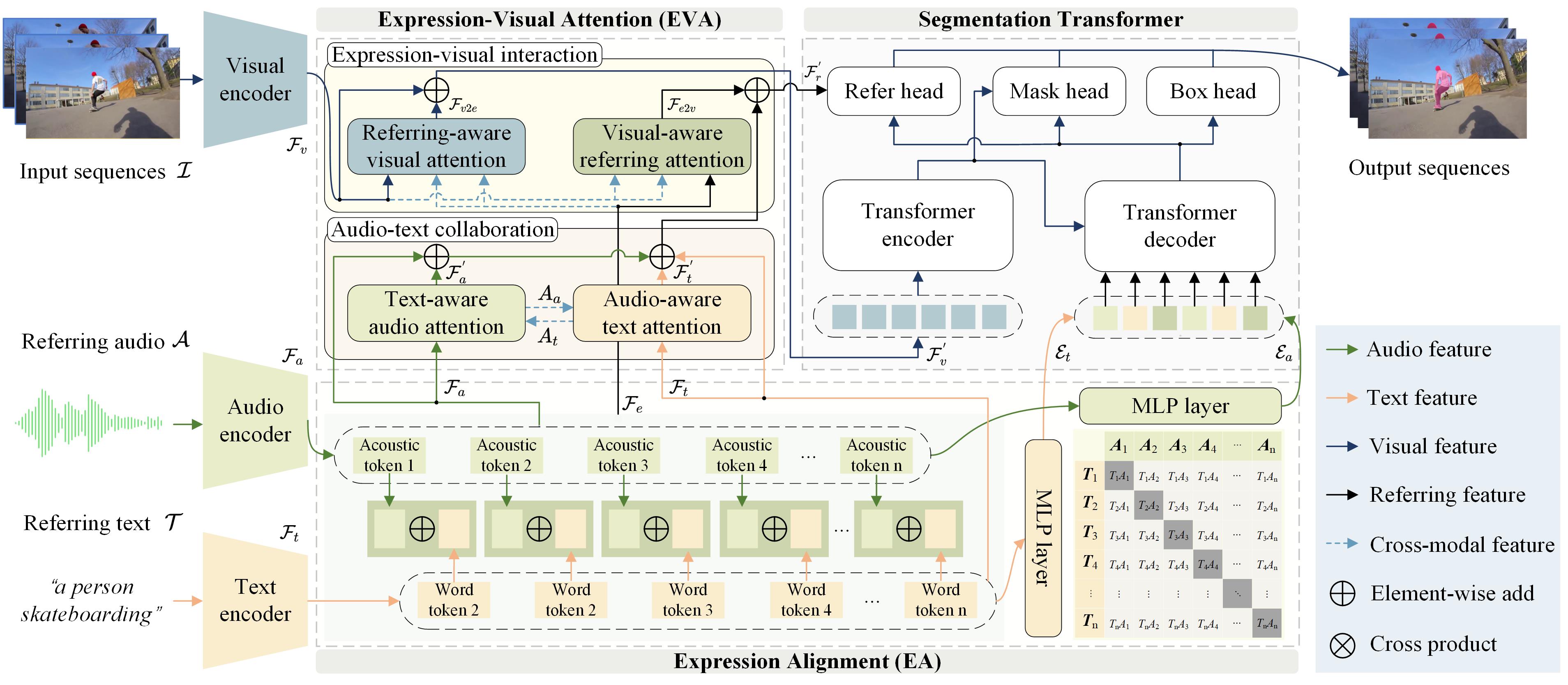}
%<*fig-framework>
\caption{\revised{An illustration of the proposed EPCFormer. Its components consist of four key stages: (1) multi-modal encoding, (2) expression alignment, (3) multi-modal interaction, and (4) segmentation and optimization.}}
%</fig-framework>
\label{fig:framework}
\end{figure*}

\section{Method} \label{sec:method}

\subsection{Overview} \label{sec:overview}

The overview of the proposed ECPFormer is shown in \revised{Figure}~\ref{fig:framework}, which consists of four components:
1) \textbf{Multi-modal encoding} refers to extracting individual feature embeddings from audio, text prompts, and video sequences. 
2) \textbf{Expression alignment} is to align the semantic-level representations of audio and text prompts.
3) \textbf{Multi-modal interaction} is to achieve comprehensive and deep fusion among the three modalities, \textit{i.e.}, audio, text, and visual features.
4) \textbf{Segmentation and optimization} refers to feeding the obtained multi-modal features into a segmentation network to generate the masks.

\subsection{Multi-modal Encoding}
\label{sec:encoding}

\textbf{Visual encoder.} 
%<*r12-i>
\revised{Given a video $\mathcal{I}\in \mathbb{R}^{T\times 3\times H \times W}$ with $T$ frames}, we investigate two main visual backbones, namely ResNet-50~\cite{ResNet} and ViT-Huge~\cite{ViT}, to extract visual features $\mathcal{F}_v\in \mathbb{R}^{C\times L_v}$ for each frame, \revised{where $H$ and $W$ are height and width of raw frame, $C$ represents the embedding dimension for Transformers, and $L_v$ represents the sequence length of flattened visual features.}
%</r12-i>

\textbf{Text encoder.} Given a text prompt $\mathcal{T}\in \mathbb{R}^N$ with $N$ words, \revised{BERT~\cite{BERT} is selected as text encoder  following~\cite{MTTR,ReferFormer,UNINEXT} to extract referring text features $\mathcal{F}_t\in\mathbb{R}^{C\times  L_t}$, where $L_t$ denotes the length of text features}.

\textbf{Audio encoder.} 
\revised{Existing methods, such as Wnet~\cite{pan2022wnet}, have difficulties extracting audio features and aligning them with visual features.} Specifically, during the process of extracting raw acoustic features following~\cite{bouchakour2018mfccs}, the network fails to extract the necessary \revised{and critical} features in a learnable manner.
\revised{The network's induction biases and imprecise object localization results are due to the indiscriminate input of all audio information, including noise.~\cite{pan2022wnet}}
Accordingly, given a reference audio $\mathcal{A}\in \mathbb{R}^{S}$ with $S$ samples, we extend the Transformer-based HuBERT~\cite{hsu2021hubert} with the designed shallow layers to extract hidden units acoustic embeddings $\mathcal{F}_a\in\mathbb{R}^{C\times L_a}$, \revised{where $L_a$ denotes the length of the audio features}.
On the one hand, unifying the features dimensions $C$ of different categories of expressions facilitates subsequent joint processing of both modalities. 
On the other hand, by fine-tuning the learnable, pre-trained backbone, we efficiently extract crucial features from the audio cues.

\subsection{Expression Alignment} \label{sec:alignment}
%<*r12-i-2>
\revised{The challenge of achieving complementarity between audio and text modalities requires alignment of inherent disparities.} However, the presence of similar and diverse expressions brings significant challenges.
\revised{Given a video frame $I \in \mathcal{I}$, there are $N_O$ objects $\{\mathcal{O}_1,\mathcal{O}_2,\dots,\mathcal{O}_{N_O}\}$ present. Any object $\mathcal{O}_i$ can be referred to by $N_T$ different text references $\{\mathcal{T}_{i,1},\mathcal{T}_{i,2},\dots,\mathcal{T}_{i,N_T}\}$ and $N_A$ different audio references $\{\mathcal{A}_{i,1},\mathcal{A}_{i,2},\dots,\mathcal{A}_{i,N_A}\}$, where $i$ is the index of the object.
We can obtain a relational mapping function, denoted as $\mathrm{Seg}(I,\mathcal{T}_{i,j})=\mathcal{O}_i$ and $\mathrm{Seg}(I,\mathcal{A}_{i,k})=\mathcal{O}_i$, where $j$ and $k$ denote $j^{th}$ text prompt and $k^{th}$ audio prompt.} 
%</r12-i-2>
As long as the textual and auditory prompts share the same meaning, both unambiguously refer to the same object, and consequently, the generated masks should be identical. 
\revised{Hence, it is crucial to ensure that text and audio features with the same meaning exhibit high similarity in the representation space. In this way,} the model can recognize that distinct types of expressions denoting the same semantics can refer to the same object.

\textbf{Expression contrastive learning.}
\revised{Different from existing methods for training batch construction~\cite{vlt,CLIP,CLAP}, we employ the most similar expressions as negative samples to increase the difficulty of contrastive learning.} 
Concretely, in each training batch, for any object $O_i, i\in \{1,2,\cdots,N_O\}$ in a video frame $I$, we randomly sample two sets of expressions from different modalities but sharing the same description for the same referred object, \revised{ denoted as $\langle \mathcal{T}_{i,j_1}, \mathcal{A}_{i,k_1}\rangle$ and $\langle \mathcal{T}_{i,j_2}, \mathcal{A}_{i,k_2}\rangle$}, where $j_1,j_2\in \{1,2,\cdots,N_T\}$ and $k_1,k_2\in \{1,2,\cdots,N_A\}$. The remaining part of the batch involves randomly selecting other video frames and corresponding text and audio.

\begin{figure*}[t!]
\centering
\includegraphics[width=1.0\linewidth]{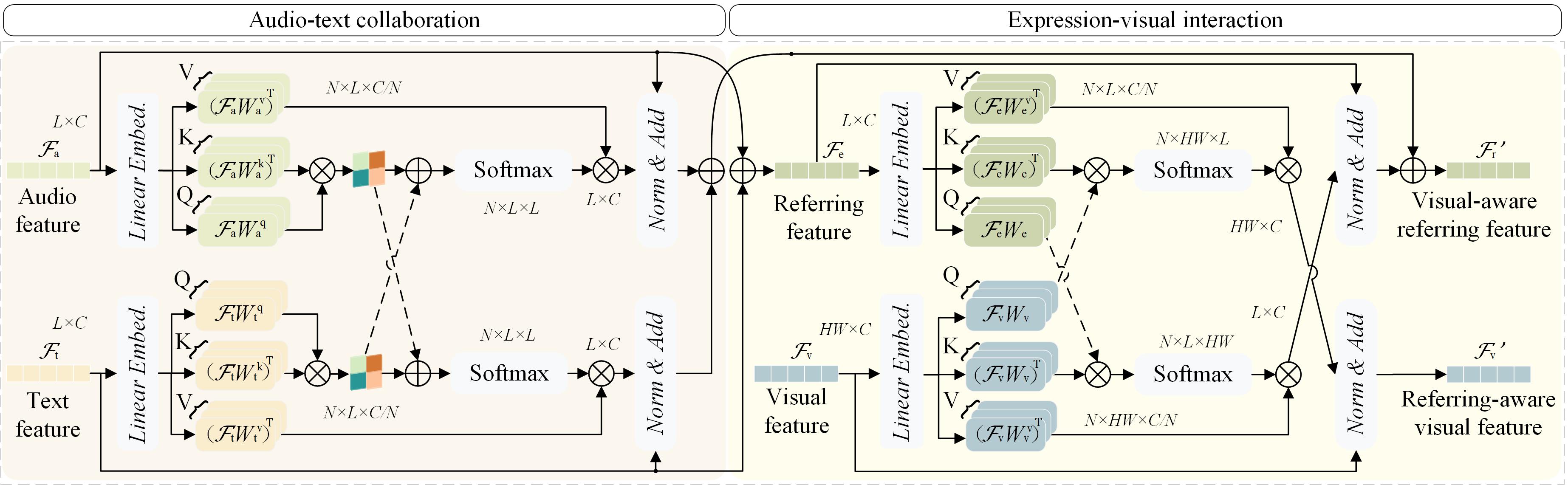}
\caption{The architecture of the proposed Expression-Visual Attention (EVA) module. EVA is a two-stream structure, comprising Audio-Text Collaboration (ATC) and Expression-Visual Interaction (EVI).
Given text, audio, and visual features obtained through their respective encoders, audio features, and text features are firstly integrated through linear combination, resulting in referring features. 
Meanwhile, the proposed ATC exploits the complementarity between text and audio features, thus facilitating the extraction of pivotal expression features.
The EVI enhances visual features by incorporating referring features and, reciprocally, refines referring features through visual features.}
\label{fig:attention_1}
\end{figure*}

During the batch mentioned above, the number of samples for one modality of expression is denoted as $N$. 
%<*r12-i-3>
\revised{First, we project audio features $\mathcal{F}_{a}$ and text features $\mathcal{F}_{t}$} into a multi-modal embedding space using a linear mapping layer, denoted as follows:
%</r12-i-3>
\begin{equation}\label{c_linear}
\begin{aligned}
\mathcal{E}_{t}&=\mathrm{MLP}(\mathrm{GAP}(\mathcal{F}_{t})),\\
\mathcal{E}_{a}&=\mathrm{MLP}(\mathrm{GAP}(\mathcal{F}_{a})),
\end{aligned}
\end{equation}
where $\mathrm{MLP}(\ast)$ denotes a multi-layer perceptron comprising two linear layers with ReLU in between.
$\mathrm{GAP}(\ast)$ denotes global average pooling.
$\mathcal{E}_{t}\in \mathbb{R}^C$ and $\mathcal{E}_{a}\in \mathbb{R}^C$ represent the obtained multi-modal embeddings, both having the same dimension $C$.
Inspired by~\cite{CLIP,CLAP}, our expression contrastive loss as follows:
%<*r12-i-4>
\begin{equation}\label{c_loss}
\begin{aligned}
\mathcal{L}_{expr}=\frac{1}{2N}\sum^N_{m=1}[&\mathrm{log}\frac{\mathrm{exp}(\mathcal{E}_{a,m} \cdot \mathcal{E}_{t,m} / \tau)}{\sum^N_{n=1}\mathrm{exp}(\mathcal{E}_{a,m} \cdot \mathcal{E}_{t,n}/\tau)} \\
+&\mathrm{log}\frac{\mathrm{exp}(\mathcal{E}_{t,m} \cdot \mathcal{E}_{a,m}/\tau)}{\sum^N_{n=1}\mathrm{exp}(\mathcal{E}_{t,m} \cdot \mathcal{E}_{a,n}/\tau)}].
\end{aligned}
\end{equation}
\revised{where $m,n\in\{1,2,\cdots,N\}$ denote the index of expression references} and $\tau$ denotes the temperature constant.
%</r12-i-4>

\revised{In this mechanism, the positive samples for one modality in each batch consist of the same semantic representation in another modality, referring to the same object in the same video frame. The negative samples encompass not only the expressions from another modality of different videos but also different semantic representations referring to the same object in the video.} In other words, this loss function compels one modality expression in the representation space to be closer to another with the same semantic content that refers to the same object in the video frame while being farther away from another modality expression with other semantic content.

%<*r26-eq>
\textbf{Expression as query.} 
\revised{The audio embeddings $\mathcal{E}_a$ can be determined best to match the text embeddings $\mathcal{E}_t$ through the cosine similarity function and vice versa.}
Unlike the query generation paradigm proposed in~\cite{ReferFormer}, the proposed Expression as Query (EQ) strategy incorporates these embeddings into the original input query embeddings of the Transformer decoder. 
%</r26-eq>
In this manner, even when the decoder receives uni-modal queries, the aligned embeddings, which can approximately represent information from another modality, provide complementary support for predicting more accurate object masks.

\subsection{Multi-modal Interaction}
\label{sec:interaction}

\revised{As aforementioned, most previous text- or audio-guided methods~\cite{pan2022wnet,MTTR,ReferFormer, vlt} concentrate on establishing relationships between video and single modality referring expression.}
Due to the inflexibility of text interactions in specific scenarios and the noise factors in audio interactions, these methods encounter limitations when deployed in practical applications. 
%To address this issue, we propose designing a unified video segmentation model for audio and text-referring expressions. Inspired by the spirits of previous works~\cite{pan2022wnet,ReferFormer}, we acknowledge that deep multi-modal fusion and interaction are crucial for developing a performant referring model. 
To facilitate effective interactions among audio, text, and vision, we further propose a \revised{Expression-Visual Attention (EVA) module}.
As illustrated in \revised{Figure}~\ref{fig:attention_1}, \revised{EVA} comprises two parallel streams, denoted as Audio-Text Collaboration (ATC) and Expression-Visual Interaction (EVI), respectively.
\revised{ATC exploits the complementarity between audio and text, refining one modality feature relative to another and vice versa.}
EVI aims to establish effective interactions between auditory and textual cues with visual features, thereby emphasizing the matching visual regions and crucial cue elements.

\textbf{Audio-text collaboration.} \revised{ATC is comprised of text-aware audio attention and audio-aware text attention, facilitating collaboration between the audio features $\mathcal{F}_a$ and the text features $\mathcal{F}_t$.}
First, we perform a linear embedding to project the features and a multiplication for each modality to obtain attention matrices $A_t$ and $A_a$ independently:
\begin{equation}\label{a_self}
\begin{aligned}
&A_{a}=\frac{\mathcal{F}_{a}W_{a}^q(\mathcal{F}_{a}W_{a}^k)^T}{\sqrt{d_k}},  \\
&A_{t}=\frac{\mathcal{F}_{t}W_{t}^q(\mathcal{F}_{t}W_{t}^k)^T}{\sqrt{d_k}},
\end{aligned}
\end{equation}
where $W_{a}^q$, $W_{a}^k$, $W_{t}^q$, and $W_{t}^q$ are the learnable linear projection matrices for the features of each modality. 
After that, we perform an addition between $A_{a}$ and $A_{t}$ to obtain a shared attention matrix $A_{e}$:
\begin{equation}\label{a_share}
\begin{aligned}
A_{e}=A_{a}+A_{t},
\end{aligned}
\end{equation}
In Eq. \ref{a_share}, when there is only a single referring modality input, either $A_{e}=A_{a}$ or $A_{e}=A_{t}$.
Consequently, ATC can still function effectively under single-modality input scenarios.
In contrast to the self-attention mechanism~\cite{vaswani2017attention}, the attention weight matrix of ATC is jointly learned from both referring modalities.
This facilitates effective exploitation of the complementarity between audio and text and enhances the capacity to extract key information from the prompting expressions.

By utilizing shared attention calculations, 
we can obtain self-attention matrices for both the audio and text modalities, enabling their mutual complementary interaction. The shared attention matrix $A_{e}$ is then used to reform an audio feature $\mathcal{F}_{a}^{'}$ and a text feature $\mathcal{F}_{t}^{'}$: 
\begin{equation}\label{a_share_soft}
\begin{aligned}
\mathcal{F}_{a}^{'}&=\mathrm{Softmax}(A_{e})(\mathcal{F}_{a}W_{a}^v)^T,\\
\mathcal{F}_{t}^{'}&=\mathrm{Softmax}(A_{e})(\mathcal{F}_{t}W_{t}^v)^T,
\end{aligned}
\end{equation}
\revised{where $W_{a}^v$ and $W_{t}^v$ are learnable linear projection matrices for the corresponding modality's features.}

\textbf{Expression-visual interaction.} \revised{Given the visual features $\mathcal{F}_v$ of the current frame, 
the acoustic features $\mathcal{F}_a$ of the audio and the linguistic features $\mathcal{F}_t$ of the sentence,}
we perform a linear combination between $\mathcal{F}_a$ and $\mathcal{F}_t$ to obtain the blended representation of the referring cues, denoted as $\mathcal{F}_e$:
\begin{equation}\label{a_add}
\begin{aligned}
\mathcal{F}_{e}=\mathcal{F}_{a}+\mathcal{F}_{t},
\end{aligned}
\end{equation}
In this way, both modalities can be placed in an equal position.
\revised{In cases where only one modality expression is available, we initialize the other modality with zero matching dimension vectors.
As a result, $\mathcal{F}_{e}$ are reduced to a single referring modality. This linearly decoupled property allows the model to process a single modality-referring expression independently.}

Afterward, inspired by~\cite{vaswani2017attention,GLIP}, referring-aware visual attention and visual-aware referring attention are employed to facilitate bi-directional interaction between referring features and visual features. Exactly, our method enables concurrent calculation of attention between text and visual, as well as between audio and visual, \revised{as} depicted as follows:
\begin{equation}\label{a_soft}
\begin{aligned}
\mathcal{F}_{v2e}&=\mathrm{Softmax}(\frac{\mathcal{F}_{v} W_{v}(\mathcal{F}_{e}W_{e})^T}{\sqrt{d_k}})(\mathcal{F}_{e}W_{e}^v)^T, \\
\mathcal{F}_{e2v}&=\mathrm{Softmax}(\frac{\mathcal{F}_{e}W_{e}(\mathcal{F}_{v}W_{v})^T}{\sqrt{d_k}})(\mathcal{F}_{v}W_{v}^v)^T,
\end{aligned}
\end{equation}
where $W_{v}$, $W_{e}$, $W_{v}^v$, and $W_{e}^v$ are learnable linear projection matrices for both features following \cite{vaswani2017attention}. After the cross-modal interaction, we perform a residual operation between the raw features and the obtained features: 
\begin{equation}\label{a_residual}
\begin{aligned}
\mathcal{F}_{v}^{'}&=\mathcal{F}_{v}+\mathcal{F}_{v2e}, \\
\mathcal{F}_{e}^{'}&=\mathcal{F}_{e}+\mathcal{F}_{e2v}.
\end{aligned}
\end{equation}

Finally, we concatenate the referring features from both branches as the output: 
\begin{equation}\label{a_output}
\begin{aligned}
\mathcal{F}_{r}^{'}=\mathcal{F}_{e}^{'}+\mathcal{F}_{a}^{'}+\mathcal{F}_{t}^{'}.
\end{aligned}
\end{equation}

In summary, the visual features gain referring awareness, and the referring features acquire visual awareness. 
In particular, when there is only a single referring expression input, either $\mathcal{F}_{e}=\mathcal{F}_{a}$ or $\mathcal{F}_{e}=\mathcal{F}_{t}$. 
By employing fixed referring expression linear projection matrices $W_{e}$ and $W_{e}^v$, \revised{EPCFormer} can seamlessly handle either the audio or text referring modality. \revised{EVA module} serves two main purposes: (1) effectively capturing the common aspects of A-VOS and R-VOS, resulting in a more generalized ability for video object segmentation according to referring expressions, and (2) alleviating overfitting when the model is in a single referring modality.

\subsection{Segmentation and Optimization} \label{subsec:segmentation_optimization}

\textbf{Segmentation Transformer.}
Following~\cite{MTTR, ReferFormer, wu2023online, UNINEXT}, the advanced Transformer is adopted as the fundamental framework for video segmentation.
Following~\cite{IDOL, GLIP, UNINEXT}, the contrastive learning loss is employed to associate each frame with the instance objects proposed by SimOTA~\cite{YOLOX}.
During the inference phase, we apply non-maximum suppression to suppress redundant candidate targets.
To distinguish between referred and non-referred objects, we compute the instance-referred matching scores, denoted as $S_{ref}$, by calculating the dot product between the instance features $\mathcal{F}_{ins}^{'}$ obtained from the decoder's output and the referring features $\mathcal{F}_{r}^{'}$ after global average pooling, \textit{i.e.,} $S_{ref}=\mathcal{F}_{ins} \mathrm{GAP}(\mathcal{F}_{r}^{'})^T$.
Following~\cite{ReferFormer, UNINEXT}, to predict high-quality masks, a dynamic convolution-based mask head~\cite{CondInst} is adopted.

\textbf{Multi-task training.}
To endow the model with the knowledge of tackling R-VOS and A-VOS concurrently during training, we propose a novel approach for multi-task joint training of both tasks.
First, we sample pairs of referring expressions with the same semantic meaning while exhibiting different modalities.
Afterward, we input the encoded alignment features between text and audio cues into the network.
Meanwhile, to avoid overfitting \revised{because} both modalities are available, we employ an equal probability dropout on either the text or audio features.
Therefore, during training, the model encounters the three tasks with equal probability: text-guided segmentation, audio-guided segmentation, and segmentation guided by both text and audio.

\textbf{Loss functions.} 
Following previous works~\cite{ReferFormer,DETR,IDOL,UNINEXT}, we adopt the following loss function to supervise the proposed model in an end-to-end manner:
\begin{equation}\label{a_loss}
\begin{aligned}
\mathcal{L}=\lambda_{ref}\mathcal{L}_{ref}+\lambda_{box}\mathcal{L}_{box}+\lambda_{mask}\mathcal{L}_{mask} \\
+\lambda_{emb}\mathcal{L}_{emb}+\lambda_{expr}\mathcal{L}_{expr},
\end{aligned}
\end{equation}
where $\mathcal{L}_{ref}$ is focal loss~\cite{lin2017focal} to classify referred and non-referred objects.
$\mathcal{L}_{box}=\mathcal{L}_{bbox}+\mathcal{L}_{giou}$ represents box regression loss, where $\mathcal{L}_{bbox}$ is $\ell_1$ loss~\cite{FasterRCNN} and $\mathcal{L}_{giou}$ is GIoU loss~\cite{GIoULoss}. 
$\mathcal{L}_{mask}=\mathcal{L}_{afl}+\mathcal{L}_{dice}$ represents mask segmentation loss, where $\mathcal{L}_{afl}$ is adaptive focal loss~\cite{lin2023adaptiveclick} and $\mathcal{L}_{dice}$ is dice loss~\cite{DiceLoss}.
$\mathcal{L}_{emb}$ is contrastive loss~\cite{IDOL} to supervise the instance embeddings across frames,
while $\mathcal{L}_{expr}$ is the proposed expression contrastive loss mentioned in \revised{Section}~\ref{sec:alignment}.
$\lambda_{ref}$, $\lambda_{box}$, $\lambda_{mask}$, $\lambda_{emb}$, and $\lambda_{expr}$ denote loss weights.

\section{Experiments} \label{sec:experiments}

\subsection{Datasets}
We conduct experiments on three datasets for R-VOS and four datasets for A-VOS, detailed as follows: 

1) Ref-Youtube-VOS~\cite{URVOS}: It is a large-scale dataset tailored for R-VOS. It encompasses $3,673$ videos with $15$K text clues for training and validation.

2) A2D-Sentences~\cite{GavrilyukGLS18}: This is created by augmenting the A2D dataset with additional textual prompt. It comprises $3,754$ videos with a collection of $6,655$ sentences.

3) J-HMDB-Sentences~\cite{GavrilyukGLS18}: It is an expansion of the J-HMDB dataset, similar to A2D-Sentences. It contains $928$ videos and their corresponding prompts.

4) Audio-Guided-VOS~\cite{pan2022wnet}: Tailored for A-VOS, this dataset is an extension that complements Ref-Youtube-VOS, A2D-Sentences, and J-HMDB-Sentences with additional $18,811$ audio prompts. 

5) A-Youtube-VOS~\cite{pan2022wnet}: This dataset is a part of Audio-Guided-VOS, encompassing a total of $11,226$ audio clues. 
Following~\cite{pan2022wnet}, the training set of Ref-Youtube-VOS is divided for building this dataset.

6) A-A2D~\cite{pan2022wnet}: It is a part of Audio-Guided-VOS, including $6,656$ audio clues. 
We denote this dataset as A-A2D to distinguish it from A2D-Sentences.

7) A-J-HMDB~\cite{pan2022wnet}: It is a part of Audio-Guided-VOS, including $928$ audio prompts. We denote this dataset as A-J-HMDB to distinguish it from J-HMDB-Sentences.

\subsection{Evaluation Metrics}
Following previous works~\cite{pan2022wnet,ReferFormer}, region similarity $\mathcal{J}$, contour accuracy $\mathcal{F}$ and their average value $\mathcal{J\&F}$ are employed to evaluate the methods on Ref-Youtube-VOS~\cite{URVOS}, Audio-Guided-VOS~\cite{pan2022wnet}, A-Youtube-VOS~\cite{pan2022wnet}, A-A2D~\cite{pan2022wnet}, and A-J-HMDB~\cite{pan2022wnet}. 
\revised{On Ref-Youtube-VOS, we upload the predictions to challenge the official server for evaluation.}
For A2D-Sentences~\cite{GavrilyukGLS18} and J-HMDB-Sentences~\cite{GavrilyukGLS18}, the Overall IoU, Mean IoU, and Precision@K, where K$\,\in\!\![0.5, 0.6, 0.7, 0.8, 0.9]$, are adopted as the evaluation metrics.

\begin{table*} [t!]
\centering
%<*tab-avos>
\renewcommand\arraystretch{1.0}
\setlength\tabcolsep{5.5pt}
\caption{Comparison in $\mathcal{J\&F}$, $\mathcal{J}$, and $\mathcal{F}$ between EPCFormer and state-of-the-art methods on Audio-Guided-VOS~\cite{pan2022wnet}, A-Youtube-VOS~\cite{pan2022wnet}, A-A2D~\cite{pan2022wnet}, and A-J-HMDB~\cite{pan2022wnet}. A-J-HMDB~\cite{pan2022wnet} is only used to evaluate the checkpoint trained on A-A2D~\cite{pan2022wnet}.  The best results are marked in \textbf{bold}, and the second-best results are \underline{underlined}.}
\resizebox{\linewidth}{!}{
\begin{tabular}{l|c|c|ccc|ccc|ccc|ccc}
\toprule [2pt]
        \multirow{2}{*}{Method} & \multirow{2}{*}{\shortstack{Visual\\Backbone}}  & \multirow{2}{*}{\shortstack{Audio\\Backbone}} & \multicolumn{3}{c|}{Audio-Guided-VOS} & \multicolumn{3}{c|}{A-Youtube-VOS} & \multicolumn{3}{c|}{A-A2D} & \multicolumn{3}{c}{A-J-HMDB}   \\ 
\cline{4-15}
\specialrule{0em}{1pt}{1pt}
        ~ & ~ & & $\mathcal{J\&F}$ & $\mathcal{J}$ & $\mathcal{F}$ & $\mathcal{J\&F}$ & $\mathcal{J}$ & $\mathcal{F}$ & $\mathcal{J\&F}$ & $\mathcal{J}$ & $\mathcal{F}$ & $\mathcal{J\&F}$ & $\mathcal{J}$ & $\mathcal{F}$  \\ 
\toprule [1pt]
URVOS+~\cite{URVOS} $_{\mathrm{ECCV2020}}$ & ResNet-50 & MFCC  & $38.2$ & $37.1$ & $39.2$ & - & - & - & - & - & - & - & - & -  \\ 
RAM+~\cite{prpe} $_{\mathrm{IJCAI2020}} $& I3D & MFCC  & $38.8$  & $38.6$  & $38.9$  & - & - & - & - & - & - & - & - & -  \\ 
VisTR+~\cite{VISTR} $_{\mathrm{CVPR2021}}$ & ResNet-50 & MFCC  & $38.8$  & $38.0$  & $39.5$  & - & - & - & - & - & - & - & - & -  \\ 
Wnet~\cite{pan2022wnet} $_{\mathrm{CVPR2022}}$  & ResNet-50 & MFCC  & $44.0$  & $43.0$  & $45.0$  & $43.6$  & $43.0$  & $44.1$  & $52.5$  & $49.8$  & $55.1$  & $61.2$  & $65.6$  & $56.7$   \\ 
\rowcolor[gray]{.9} \textbf{EPCFormer} (ours) & ResNet-50 & HuBERT  & \underline{$54.3$} & \underline{$54.3$} & \underline{$54.2$} & \underline{$53.7$}  & \underline{$52.4$}  & \underline{$55.0$} & \underline{$63.0$} &	\underline{$60.7$} &	\underline{$65.2$} &	\underline{$62.6$} &	\underline{$67.4$} &	\underline{$57.9$}    \\ 

\rowcolor[gray]{.9} \textbf{EPCFormer} (ours) & ViT-H & HuBERT  & $\mathbf{59.0}$ & $\mathbf{58.9}$ & $\mathbf{59.1}$ & $\mathbf{56.7}$ & $\mathbf{55.0}$ & $\mathbf{58.5}$ & $\mathbf{64.9}$ &	$\mathbf{62.6}$ &	$\mathbf{67.3}$  & $\mathbf{63.7}$ & $\mathbf{68.5}$ &  $\mathbf{58.8}$ \\ 
\toprule [2pt]
\end{tabular}
}
%</tab-avos>
\label{tab:avos}
\end{table*}

\begin{table*} [t!]
\centering
%<*tab-rvos-a2d>
\renewcommand\arraystretch{1.0}
\setlength\tabcolsep{6pt}
\caption{Comparison in Precision@K, Overall IoU, and Mean IoU between EPCFormer and state-of-the-art methods on A2D-Sentences~\cite{GavrilyukGLS18}.}
\resizebox{\linewidth}{!}{
\begin{tabular}{l|
c|c|cccccccc}
\toprule [2pt]
        \multirow{2}{*}{Method} & \multirow{2}{*}{\shortstack{Visual\\Backbone}}  & \multirow{2}{*}{\shortstack{Text\\Backbone}}  & \multicolumn{5}{c}{Precision}  & \multicolumn{2}{c}{IoU} & \multirow{2}{*}{mAP}  \\ 
\cline{4-10}
\specialrule{0em}{1pt}{1pt}
        ~ & ~ ~ & & P@0.5 & P@0.6 & P@0.7 & P@0.8 & P@0.9 & Overall & Mean  \\ 
\toprule [1pt]
        ACAN~\cite{wang2019acan} $_{\mathrm{ICCV2019}}$ & I3D & Word2Vec & $55.7$  & $45.9$  & $31.9$  & $16.0$  & $2.0$  & $60.1$  & $49.0$ & $27.4$     \\ 
        CMSA + CFSA~\cite{ye2021referring} $_{\mathrm{TPAMI2022}}$ & ResNet-101 & - & $48.7$  & $43.1$  & $35.8$  & $23.1$  & $5.2$  & $61.8$  & $43.2$ & -   \\ 
        CSTM~\cite{hui2021collaborative}  $_{\mathrm{CVPR2021}}$ & I3D & GRU & $65.4$  & $58.9$  & $49.7$  & $33.3$  & $9.1$  & $66.2$  & $56.1$  & $39.9$    \\ 
        CMPC-V~\cite{liu2022cmpc} $_{\mathrm{TPAMI2022}}$ & I3D & LSTM  & $65.5$  & $59.2$  & $50.6$  & $34.2$  & $9.8$  & $65.3$  & $57.3$  & \underline{$40.4$}   \\ 
        ClawCraneNet~\cite{liang2021clawcranenet} $_{\mathrm{Arxiv2021}}$ & ResNet-50 &  bi-LSTM  & \underline{$70.4$}  & \underline{$67.7$}  & \underline{$61.7$}  & \underline{$48.9$}  & \underline{$17.1$}  & \underline{$63.1$}  & \underline{$59.9$} & -     \\ 
        \rowcolor[gray]{.9}\textbf{EPCFormer} (ours) & ResNet-50 & BERT  & $\mathbf{80.2}$  & $\mathbf{78.1}$  & $\mathbf{72.1}$  & $\mathbf{56.4}$  & $\mathbf{20.7}$  & $\mathbf{74.6}$  & $\mathbf{67.9}$  & $\mathbf{51.7}$    \\ 
\toprule [1pt]
        MTTR~\cite{MTTR} $_{\mathrm{CVPR2022}}$ & Video-Swin-T & RoBERTa   & $75.4$  & $71.2$  & $63.8$  & $48.5$  & $16.9$  & $72.0$  & $64.0$ & $46.1$     \\ 
        ReferFormer~\cite{ReferFormer} $_{\mathrm{CVPR2022}}$ & Video-Swin-T & RoBERTa  & $82.8$  & $79.2$  & $72.3$  & $55.3$  & $19.3$  & $77.6$  & $69.6$  & $52.8$   \\ 
        ReferFormer~\cite{ReferFormer} $_{\mathrm{CVPR2022}}$ & Video-Swin-B & RoBERTa  & $83.1$  & $80.4$  & $74.1$  & $57.9$  & $21.2$  & $78.6$ & $70.3$ & $55.0$     \\ 
        % DMFormer~\cite{DMFormer} $_{\mathrm{TCSVT2023}}$ & Swin-T & BERT  & $81.3$  & $78.8$  & $71.9$  & $55.2$  & $20.3$  & $76.0$  & $68.3$  & $54.3$   \\ 
        % DMFormer~\cite{DMFormer} $_{\mathrm{TCSVT2023}}$  & Swin-L & BERT   & \underline{$83.7$}  & \underline{$81.8$}  & \underline{$75.7$}  & \underline{$60.0$}  & \underline{$24.3$}  & $78.4$  & $70.9$ & \underline{$58.2$}     \\ 
        % OnlineRefer~\cite{wu2023online} $_{\mathrm{ICCV2023}}$ & Video-Swin-B & RoBERTa  & $83.1$ & $80.2$ & $73.4$ & 56.8 & $21.7$ & $79.6$ & $70.5$ & -  \\ 
        SgMg~\cite{miao2023spectrum} $_{\mathrm{ICCV2023}}$ & Video-Swin-T & RoBERTa  & - & - & - & - & - & $78.0$ & $70.4 $& $56.1$  \\ 
        SgMg~\cite{miao2023spectrum} $_{\mathrm{ICCV2023}}$ & Video-Swin-B & RoBERTa  & - & - & - & - & - & \underline{$79.9$} & \underline{$72.0$} & $\mathbf{58.5}$  \\ 
        HTML~\cite{han2023html} $_{\mathrm{ICCV2023}}$ & Video-Swin-T & RoBERTa  & $82.2$ & $79.2$ & $72.3$ & $55.3$ & $20.1$ & $77.6$ & $69.2$ & $53.4$  \\ 
        HTML~\cite{han2023html} $_{\mathrm{ICCV2023}}$ & Video-Swin-B & RoBERTa  & \underline{$84.0$} & \underline{$81.5$} & \underline{$75.8$} & \underline{$59.2$} & \underline{$22.8$} & $79.5$ & $71.2$  & $56.7$  \\ 

        \rowcolor[gray]{.9}\textbf{EPCFormer} (ours) & ViT-H & BERT  & $\mathbf{84.6}$ &	$\mathbf{83.5}$ &	$\mathbf{78.8}$ &	$\mathbf{66.0}$ &	$\mathbf{28.1}$ &	$\mathbf{80.6}$ &	$\mathbf{72.6}$ &	\underline{$58.2$}    \\ \toprule [2pt]
\end{tabular}
} 
%</tab-rvos-a2d>
\label{tab:a2d}
\end{table*}

\begin{table*} [t!]
\centering
%<*tab-rvos-jhmdb>
\renewcommand\arraystretch{1.0}
\setlength\tabcolsep{6pt}
\caption{Comparison in Precision@K, Overall IoU, and Mean IoU between EPCFormer and state-of-the-art methods on J-HMDB-Sentences~\cite{GavrilyukGLS18}.}
\resizebox{\linewidth}{!}{
\begin{tabular}{l|c|c|cccccccc}
\toprule [2pt]
        \multirow{2}{*}{Method} & \multirow{2}{*}{\shortstack{Visual\\Backbone}}  & \multirow{2}{*}{\shortstack{Text\\Backbone}}  & \multicolumn{5}{c}{Precision}  & \multicolumn{2}{c}{IoU} & \multirow{2}{*}{mAP}   \\ 
\cline{4-10}
\specialrule{0em}{1pt}{1pt}
        ~ & ~ ~ & & P@0.5 & P@0.6 & P@0.7 & P@0.8 & P@0.9 & Overall & Mean    \\ 
\toprule [1pt]
        ACAN~\cite{wang2019acan} $_{\mathrm{ICCV2019}}$ & I3D & Word2Vec & $75.6$  & $56.4$  & $28.7$  & $3.4$  & $0.0$  & $57.6$  & $58.4$ & $28.9$     \\ 
        CMSA + CFSA~\cite{ye2021referring} $_{\mathrm{TPAMI2022}}$ & ResNet-101 & - & $76.4$  & $62.5$  & $38.9$  & $9.0$  & \underline{$0.1$}  & $62.8$ & $58.1$ & -   \\ 
        CSTM~\cite{hui2021collaborative}  $_{\mathrm{CVPR2021}}$ & I3D & GRU & $78.3$  & $63.9$  & $37.8$  & $7.6$  & $0.0$  & $59.8$  & $60.4$ & $33.5$    \\ 
        CMPC-V~\cite{liu2022cmpc} $_{\mathrm{TPAMI2022}}$ & I3D & LSTM & $81.3$  & $65.7$  & $37.1$  & $7.0$  & $0.0$  & $61.6$  & $61.7$ & \underline{$34.2$}   \\ 
        ClawCraneNet~\cite{liang2021clawcranenet}$_{\mathrm{ArXiv2021}}$ & ResNet-50 & bi-LSTM & \underline{$88.0$}  & \underline{$79.6$}  & \underline{$56.6$}  & \underline{$14.7$}  & $\mathbf{0.2}$  & \underline{$64.4$}  & \underline{$65.6$} & -     \\ 
        \rowcolor[gray]{.9}\textbf{EPCFormer} (ours) & ResNet-50 & BERT & $\mathbf{94.8}$ &	$\mathbf{89.1}$ & $\mathbf{66.7}$ &	$\mathbf{18.9}$ &	$0.0$ &	$\mathbf{71.1}$ &	$\mathbf{70.7}$ &	$\mathbf{42.8}$  \\ 
\toprule [1pt]
        MTTR~\cite{MTTR} $_{\mathrm{CVPR2022}}$ & Video-Swin-T & RoBERTa & $93.9$  & $85.2$  & $61.6$  & $16.6$  & $0.1$  & $70.1$  & $69.8$  & $39.2$  \\ 
        ReferFormer~\cite{ReferFormer} $_{\mathrm{CVPR2022}}$ & Video-Swin-T & RoBERTa & $95.8$  & $89.3$  & $66.8$  & $18.9$ & \underline{$0.2$}  & $71.9$  & $71.0$ & $42.2$    \\ 
        ReferFormer~\cite{ReferFormer} $_{\mathrm{CVPR2022}}$ & Video-Swin-B & RoBERTa & \underline{$96.2$}  & \underline{$90.2$}  & \underline{$70.2$}  & \underline{$21.0$}  & $\mathbf{0.3}$  & $73.0$  & $71.8$  & $43.0$     \\ 
        SgMg~\cite{miao2023spectrum} $_{\mathrm{ICCV2023}}$ & Video-Swin-T & RoBERTa & - & - & - & - & - & $72.8$ & $71.7$ & $44.4$  \\ 
        SgMg~\cite{miao2023spectrum} $_{\mathrm{ICCV2023}}$ & Video-Swin-B & RoBERTa & - & - & - & - & - & \underline{$73.7$} & \underline{$72.5$} & \underline{$45.0$}  \\ 
        HTML~\cite{han2023html} $_{\mathrm{ICCV2023}}$ & Video-Swin-T & RoBERTa  & - & - & - & - & - & - & -  & $42.7$  \\ 
        HTML~\cite{han2023html} $_{\mathrm{ICCV2023}}$ & Video-Swin-B & RoBERTa  & - & - & - & - & - & - & -  & $44.2$  \\ 

        \rowcolor[gray]{.9}\textbf{EPCFormer} (ours) & ViT-H & BERT & $\mathbf{97.6}$ &	$\mathbf{93.1}$ &	$\mathbf{72.6}$ &	$\mathbf{23.0}$ &	$0.0$ &	$\mathbf{74.0}$ & $\mathbf{73.1}$ &	$\mathbf{45.5}$     \\ 
\toprule [2pt]
\end{tabular}
}
%</tab-rvos-jhmdb>
\label{tab:jhmdb}
\end{table*}

\subsection{Implementation Details} \label{subsec:implementation_details}
\textbf{Training setting.} This work implements the proposed method with a frozen BERT-base~\cite{BERT} and an unfrozen HuBERT-Base~\cite{hsu2021hubert}.
Following~\cite{DeformableDETR}, the Transformer encoder and decoder are configured with $6$ layers.
The number of the Transformer decoder's queries is set to $900$.
Following~\cite{UNINEXT}, we randomly sample $2$ frames during training and only $1$ frame during inference.
The AdamW optimizer~\cite{AdamW} is adopted with an initial learning rate of $10^{-4}$ and weight decay of $0.05$.
The model is trained on two NVIDIA RTX A6000 GPUs \revised{with} $48$G \revised{of} RAM, with a batch size of $2$ and $2$ pairs of frames per GPU. 
For a fair comparison, our models are initialized by pre-trained weights following~\cite{ReferFormer,wu2023online,UNINEXT}.
Following~\cite{wu2023online}, the loss weights $\lambda_{ref}$, $\lambda_{box}$, $\lambda_{mask}$, $\lambda_{emb}$, and $\lambda_{expr}$ are set $2$, $5$, $5$, $2$, and $1$, respectively.
We conduct joint training for $150,000$ iterations on Ref-Youtube-VOS and A-Youtube-VOS, $50,000$ iterations on A2D-Sentences~\cite{GavrilyukGLS18} and A-A2D~\cite{pan2022wnet}, and $150,000$ iterations on Audio-Guided-VOS~\cite{pan2022wnet}.

\textbf{Training dataset details.}
We deploy a generalist model to handle modalities encompassing text, audio, and videos.
The training data from A-VOS and R-VOS \revised{is} concurrently collected to facilitate joint training.
More precisely, during a single forward propagation, a video, shared across both tasks, is utilized along with audio from A-VOS and text from R-VOS. 
Specifically, Ref-Youtube-VOS~\cite{URVOS} is paired with A-Youtube-VOS~\cite{pan2022wnet}, and A2D-Sentence~\cite{GavrilyukGLS18} is paired with A-A2D~\cite{pan2022wnet}. 
Diverging slightly from prior methods~\cite{MTTR,ReferFormer}, 
due to the testing set of A-Youtube-VOS~\cite{pan2022wnet} being derived from the training set of Ref-Youtube-VOS~\cite{URVOS},
we use only overlapping training data for multi-task to prevent potential data leakage in A-VOS.

\textbf{Testing dataset details.} We evaluate the proposed model on A-VOS and R-VOS with single-modality prompts. 
Additionally, we examine the model's performance in scenarios involving simultaneous text and audio prompts to explore their combined effects. 
In the evaluation of A-VOS, the testing set comprises A-Youtube-VOS~\cite{pan2022wnet}, A-A2D~\cite{pan2022wnet}, A-J-HMDB~\cite{pan2022wnet}, and Audio-Guided-VOS~\cite{pan2022wnet}, while for R-VOS, it comprises Ref-Youtube-VOS~\cite{URVOS}, A2D-Sentences~\cite{GavrilyukGLS18}, and J-HMDB-Sentences~\cite{GavrilyukGLS18}.
Specifically, for A-J-HMDB~\cite{pan2022wnet} and J-HMDB-Sentences~\cite{GavrilyukGLS18}, we directly report the results utilizing the jointly trained weights obtained from A-A2D~\cite{pan2022wnet} and A2D-Sentences~\cite{GavrilyukGLS18} without finetuning.

\subsection{Comparison Methods}
A wide variety of state-of-the-art A-VOS and R-VOS methods are incorporated for comparison:

1) A-VOS methods: The A-VOS methods include Wnet~\cite{pan2022wnet}, URVOS+~\cite{URVOS}, RAM+~\cite{prpe}, and VisTR+~\cite{VISTR}. 

2) R-VOS methods: \revised{Including CMPC-V~\cite{liu2022cmpc}, URVOS~\cite{URVOS},
YOFO~\cite{YOFO}, LBDT~\cite{LBDT}, MLSA~\cite{wu2022multi}, VLT~\cite{vlt}, MTTR~\cite{MTTR}, ReferFormer~\cite{ReferFormer}, SgMg~\cite{miao2023spectrum}, ACAN~\cite{wang2019acan},
CMSA+CFSA~\cite{ye2021referring}, CSTM~\cite{hui2021collaborative}, CMPC-V~\cite{liu2022cmpc}, 
ClawCraneNet~\cite{liang2021clawcranenet}, HTML~\cite{han2023html}, TempCD~\cite{tang2023temporal}, and
R$^2$-VOS~~\cite{li2023robust}.}

\subsection{Comparison with State-of-the-Art A-VOS Methods}
\label{subsec:avos}

Table~\ref{tab:avos} lists the results of different A-VOS methods on four datasets. \revised{The results show that EPCFormer achieves state-of-the-art performance in various scenarios and with different types of objects.} This is attributed to effectively establishing interactions and complementarity between multiple modalities. The following obvious findings can be observed: 1) Compared with the off-the-shelf methods using ResNet-50 as the backbone, EPCFormer achieves the $\mathcal{J\&F}$ of $54.3\%$, $53.7\%$, $63.0\%$, and $62.6\%$ on Audio-Guided-VOS~\cite{pan2022wnet}, A-Youtube-VOS~\cite{pan2022wnet}, A-A2D~\cite{pan2022wnet}, and A-J-HMDB~\cite{pan2022wnet}, respectively, which are $10.3\%$, $10.1\%$, $11.5\%$, and $1.4\%$ higher than Wnet~\cite{pan2022wnet}. \revised{2) With a stronger ViT-Huge backbone, EPCFormer further boosts the performance and achieves the $\mathcal{J\&F}$ of $59.0\%$, $56.7\%$, $64.9\%$, and $63.7\%$ on all datasets, respectively.} 

\begin{figure}[t!]
\centering
\includegraphics[width=1.0\linewidth]{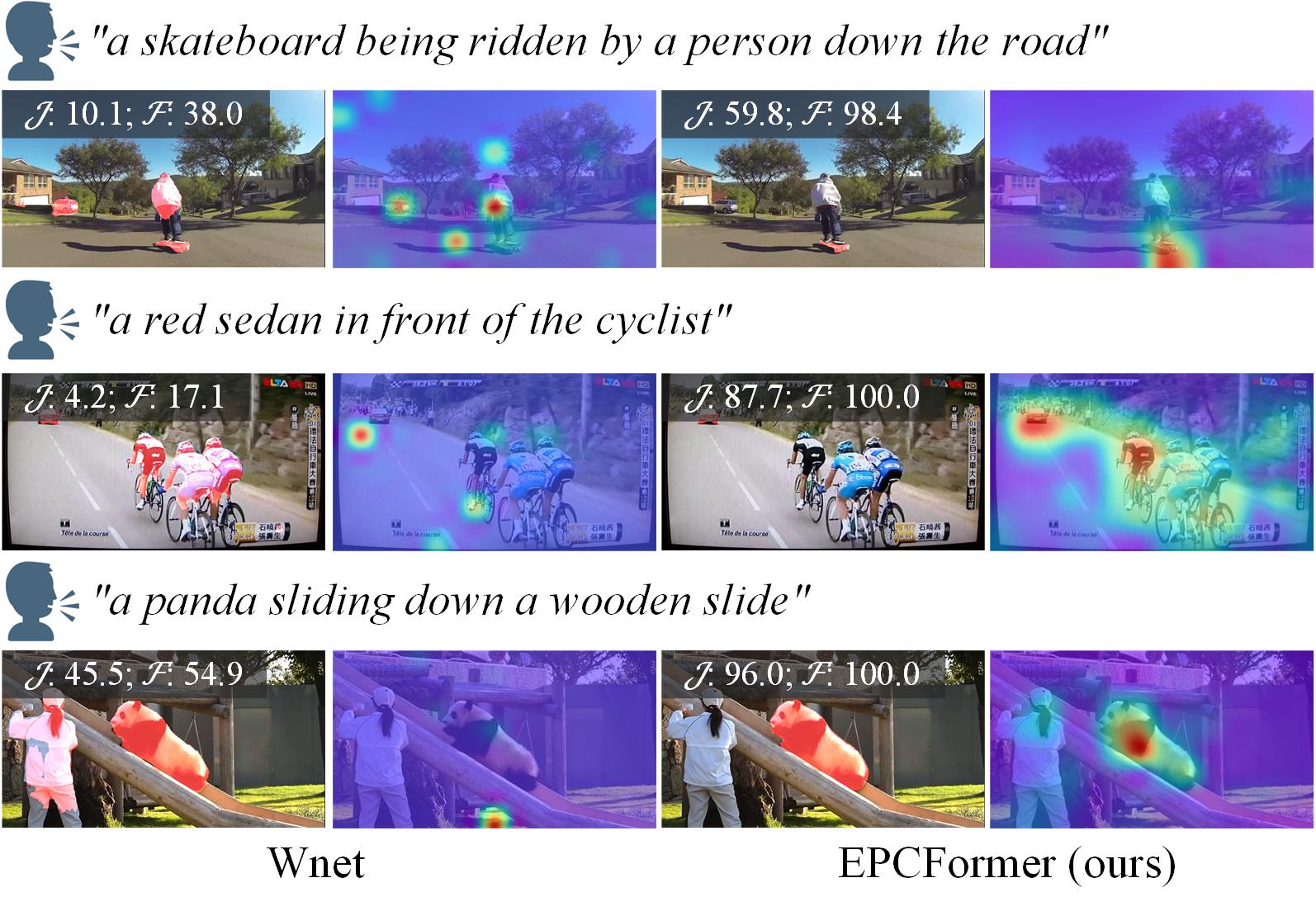}
%<*fig-avos-cam>
\caption{Visualization of heat maps in the proposed EPCFomrer and Wnet~\cite{pan2022wnet} on A-Youtube-VOS~\cite{pan2022wnet} \revised{dataset}. EPCFormer demonstrates the capability to discern referred objects based on \revised{audio prompts}.}
%</fig-avos-cam>
\label{fig:avos_cam}
\end{figure}

\begin{figure}[t!]
\centering
\includegraphics[width=0.99\linewidth]{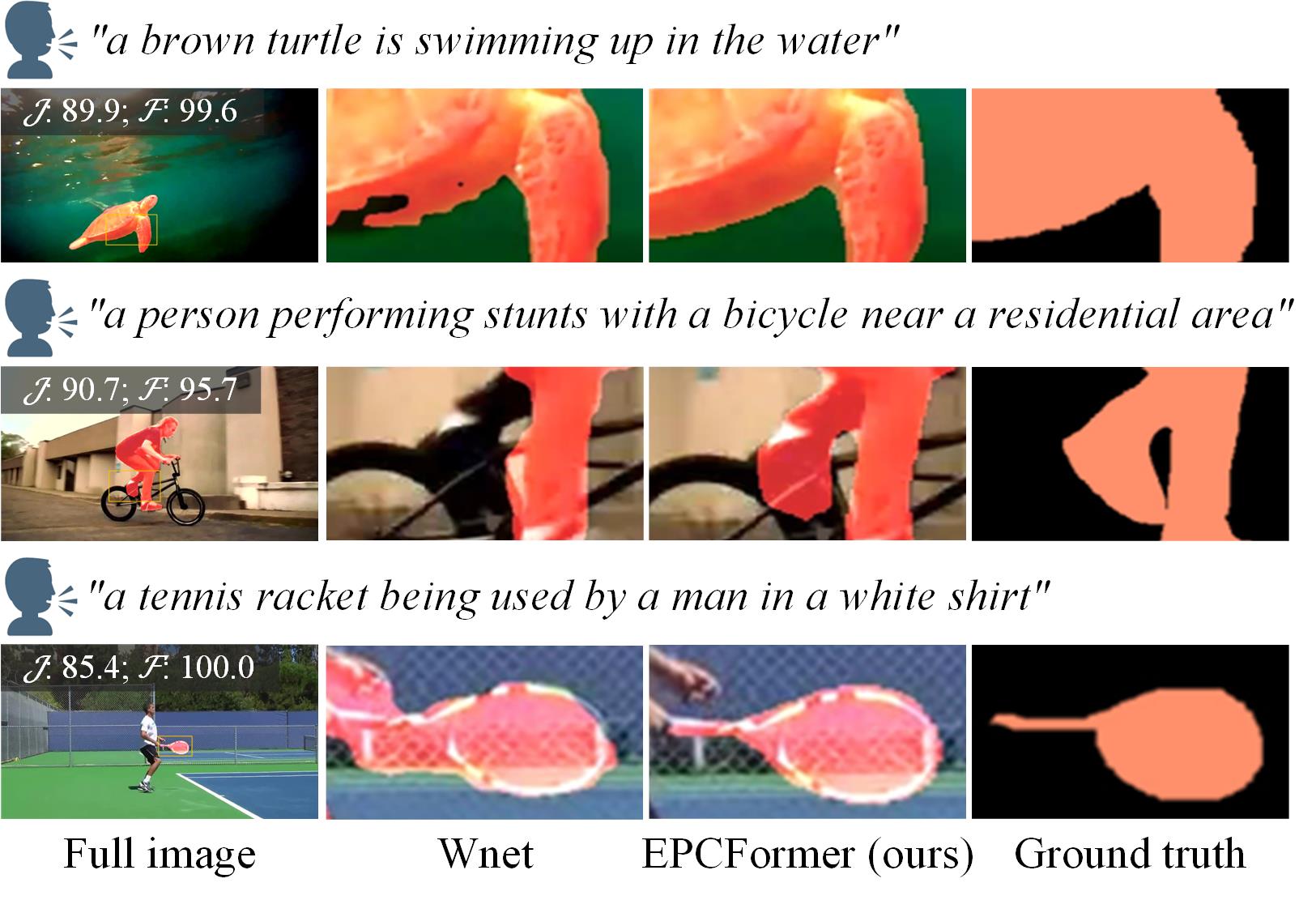}
\caption{The quality results in EPCFomrer and Wnet~\cite{pan2022wnet} on A-Youtube-VOS~\cite{pan2022wnet}. The proposed EPCFomrer can generate more detailed masks.}
\label{fig:avos_detail}
\end{figure}

\begin{figure*}[t!]
\centering
\includegraphics[width=1.0\textwidth]{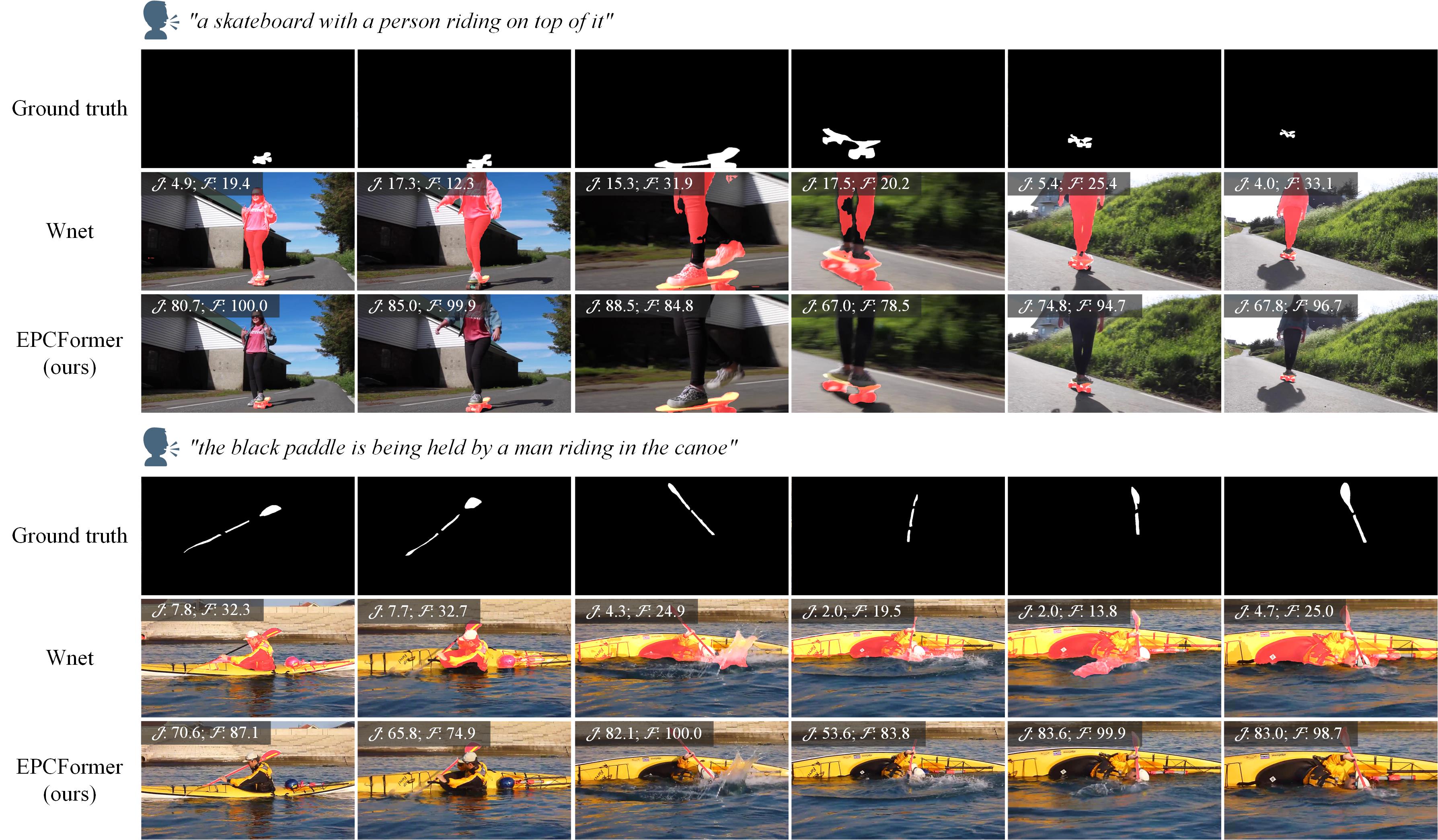}
\caption{Segmentation results of the proposed EPCFomrer and Wnet~\cite{pan2022wnet} on A-Youtube-VOS~\cite{pan2022wnet}. The segmentation maps are superimposed in orange over the original images. EPCFormer exhibits comprehensive exploitation of audio prompts, resulting in accurate localization and precise segmentation of referred objects.}
\label{fig:avos}
\end{figure*}

In addition, \revised{Figures~\ref{fig:avos_cam},~\ref{fig:avos_detail} and~\ref{fig:avos}} show the visualization results of EPCFormer and Wnet~\cite{pan2022wnet} on A-Youtube-VOS. \revised{Specifically, in Figure}~\ref{fig:avos_cam}, the heat maps generated by EPCFormer showcase the superior comprehension of the audio prompts, leading to accurate localization of the referred objects.
In the \revised{$1^{\mathrm{st}}$ example}, EPCFormer demonstrates a strong understanding of the keywords ``skateboard'', ``person'', and ``road'', as well as their relationships within the given \revised{audio prompts}.
The proposed EVA assists the model in prioritizing the skateboard, mitigating potential \revised{distractions} from the playing person.
This ability to precisely analyze referred objects is once again confirmed in \revised{Figure}~\ref{fig:avos_detail}.
In particular, the \revised{$1^{\mathrm{st}}$ example} displays that our model can predict more accurate confidence for the edge of the referred turtle, resulting in the formation of a more complete mask.
Furthermore, in \revised{Figure}~\ref{fig:avos}, the global localization capability of EPCFormer for referred objects throughout the entire video is demonstrated.

\begin{table} [t!]
\centering
\renewcommand\arraystretch{1.1}
\setlength\tabcolsep{0.5pt}
%<*tab-rvos>
\caption{Comparison in $\mathcal{J\&F}$, $\mathcal{J}$, and $\mathcal{F}$ between EPCFormer and state-of-the-art methods on Ref-Youtube-VOS~\cite{URVOS}.}
\resizebox{\linewidth}{!}{
\begin{tabular}{l|c|c|ccc}
\toprule [2pt]
        \multirow{2}{*}{Method} & \multirow{2}{*}{\shortstack{Visual\\Backbone}} & \multirow{2}{*}{\shortstack{Text\\ Backbone}} & \multicolumn{3}{c}{Ref-Youtube-VOS}   \\ 
\cline{4-6}
\specialrule{0em}{1pt}{1pt}
        ~ & ~ & ~ & $\mathcal{J\&F}$ & $\mathcal{J}$ & $\mathcal{F}$  \\ 
\toprule [1pt]
        CMPC-V~\cite{liu2022cmpc} $_{\mathrm{TPAMI2022}}$ & I3D & LSTM & $47.5$  & $45.6$  & $49.3$   \\ 
        URVOS~\cite{URVOS} $_{\mathrm{ECCV2022}}$ & ResNet-50 & - & $47.2$  & $45.3$  & $49.2$   \\ 
        YOFO~\cite{YOFO} $_{\mathrm{AAAI2022}}$ & ResNet-50 & BERT & $48.6$  & $47.5$  & $49.7$   \\ 
        LBDT~\cite{LBDT} $_{\mathrm{CVPR2022}}$ & ResNet-50 & LSTM & $49.4$  & $48.2$  & $50.6$   \\ 
        MLSA~\cite{wu2022multi} $_{\mathrm{CVPR2022}}$ & ResNet-50 & Transformer & \underline{$49.7$}  & $48.4$  & $51.0$   \\ 
        ReferFormer~\cite{ReferFormer} $_{\mathrm{CVPR2022}}$ & ResNet-50 & RoBERTa & $\mathbf{55.6}$  & $\mathbf{54.8}$  & \underline{$56.5$}   \\ 
        
        \rowcolor[gray]{.9}\textbf{EPCFormer} (ours) & ResNet-50 & BERT & $\mathbf{55.6}$  & \underline{$53.9$}  & $\mathbf{57.2}$   \\ 
\toprule [1pt]
        MTTR~\cite{MTTR} $_{\mathrm{CVPR2022}}$ & Video-Swin-T & RoBERTa & $55.3$  & $54.0$  & $56.6$   \\ 
        VLT~\cite{vlt} $_{\mathrm{TPAMI2022}}$ & Video-Swin-B & BERT & \underline{$63.8$}  & \underline{$61.9$}  & \underline{$65.6$}   \\ 
        ReferFormer~\cite{ReferFormer} $_{\mathrm{CVPR2022}}$  & Swin-L & RoBERTa & $62.4$  & $60.8$  & $64.0$   \\ 
        ReferFormer~\cite{ReferFormer} $_{\mathrm{CVPR2022}}$  & Video-Swin-B & RoBERTa & $62.9$  & $61.3$  & $64.6$   \\ 
        SgMg~\cite{miao2023spectrum} $_{\mathrm{ICCV2023}}$ & Video-Swin-T & RoBERTa & $62.0$  & $60.4$  & $63.5$   \\ 
        TempCD~\cite{tang2023temporal} $_{\mathrm{ICCV2023}}$ & Video-Swin-T & RoBERTa & $62.3$  & $60.5$  & $64.0$   \\
        R$^2$-VOS~~\cite{li2023robust} $_{\mathrm{ICCV2023}}$ & Video-Swin-T & RoBERTa & $61.3$  & $59.6$  & $63.1$   \\
        HTML~\cite{han2023html}  $_{\mathrm{ICCV2023}}$ & Swin-L & RoBERTa & $63.4$  & $61.5$  & $65.3$   \\
        HTML~\cite{han2023html}  $_{\mathrm{ICCV2023}}$ & Video-Swin-T & RoBERTa & $61.2$  & $59.5$  & $63.0$   \\
        HTML~\cite{han2023html}  $_{\mathrm{ICCV2023}}$ & Video-Swin-B& RoBERTa & $63.4$  & $61.5$  & $65.2$   \\
        
        \rowcolor[gray]{.9}\textbf{EPCFormer} (ours) & ViT-H & BERT & $\mathbf{65.0}$  & $\mathbf{62.9}$  & $\mathbf{67.2}$ \\
\toprule [2pt]
\end{tabular}
}
%</tab-rvos>
\label{tab:rvos}
\end{table}

\begin{figure*}[t!]
\centering
\includegraphics[width=1.0\textwidth]{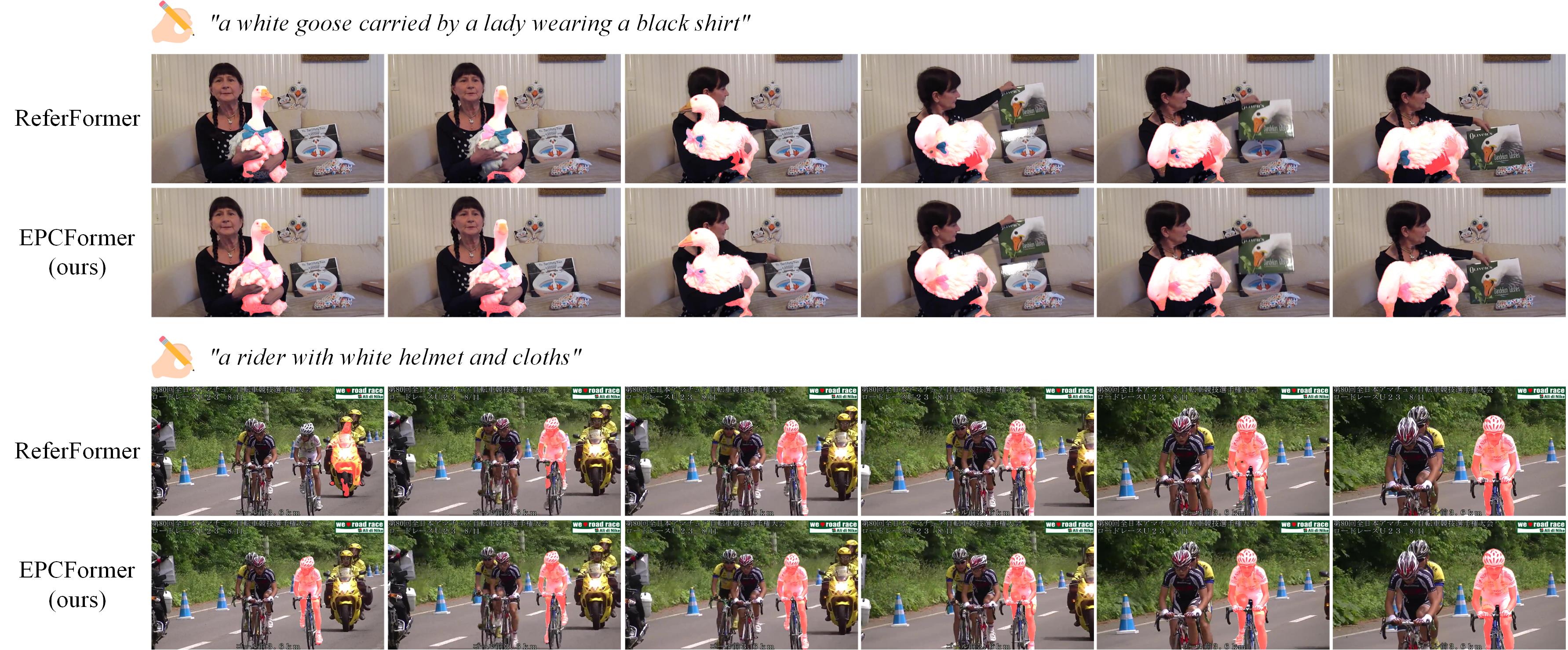}
\caption{Segmentation results of EPCFomrer and ReferFormer~\cite{ReferFormer} on Ref-Youtube-VOS~\cite{URVOS}. EPCFormer can more accurately locate specific targets and generate more detailed masks.}
\label{fig:rvos}
\end{figure*}

\begin{table*} [t!]
\centering
%<*tab-hubert>
\renewcommand\arraystretch{1.15}
\setlength\tabcolsep{2.2pt}
\caption{Model generalizability analysis on Ref-Youtube-VOS~\cite{URVOS} and A-Youtube-VOS~\cite{pan2022wnet}. 
Noting that when calculating the computational efficiency, the input image size is $480\times 853$ for all methods. Note that $^\dagger$ indicates the test input of the model is both audio and text. \revised{All experiments are conducted on 2 NVIDIA RTX A6000 GPUs with 48 GRAM.}}
\resizebox{\linewidth}{!}{
\begin{tabular}{l|c|c|c|cccccc|cccccc|c}
\toprule [2pt]
\multirow{2}{*}{Method} & \multirow{2}{*}{\shortstack{Visual\\Backbone}} & \multirow{2}{*}{\shortstack{Text\\ Backbone}} & \multirow{2}{*}{\shortstack{Audio\\Backbone}} &  \multicolumn{6}{c|}{Ref-Youtube-VOS}    & \multicolumn{6}{c|}{A-Youtube-VOS} & \revised{\multirow{2}{*}{\shortstack{Training\\Time}}} \\ 
\cline{5-16}
~ & ~ & ~ & ~ & $\mathcal{J\&F}$ & $\mathcal{J}$ & $\mathcal{F}$ & Param/M & FLOPs/G & FPS & $\mathcal{J\&F}$ & $\mathcal{J}$ & $\mathcal{F}$ & Param/M & FLOPs/G & FPS \\ 
\toprule [1pt]
Wnet~\cite{pan2022wnet} & ResNet-50 & - & MFCC  & - & - & - & - & - & - & $43.6$  & $43.0$  & $44.1$ & $38.88$ & $79.34$ & $5.25$ & \revised{$47$ h $39$ min} \\ 
Wnet+~\cite{pan2022wnet} & ResNet-50 & - & HuBERT  & - & - & - & - & - & -  & $41.9$  & $41.8$  & $42.0$ & $133.53$ & $196.85$ &  $23.25$ & \revised{$55$ h $12$ min}  \\ 
Wnet++~\cite{pan2022wnet} & ResNet-50 & BERT& -  &  $46.8$  & $46.3$ & $47.4$ & $124.81$ & $80.52$ & $25.68$  & - & -  & -  & - & - & - & \revised{$53$ h $51$ min} \\
\rowcolor[gray]{.9}\textbf{EPCFormer+} (ours) & ResNet-50 & - & MFCC  &  - & -  & - & - & - &  - & $49.8$  & $48.4$  & $51.2$ & $46.37$ & $230.06$ & $4.68$  & \revised{$70$ h $26$ min} \\ 
\rowcolor[gray]{.9}\textbf{EPCFormer} (ours) & ResNet-50  & - & HuBERT &  - & -  & - & - & - &  - & $\mathbf{53.7}$ & $\mathbf{52.4}$  & $\mathbf{55.0}$ & $143.05$ & $395.41$ & $13.75$  & \revised{$72$ h $17$ min} \\ 
\toprule [1pt]
ReferFormer~\cite{ReferFormer} & ResNet-50 & RoBERTa  & - & $\mathbf{55.6}$  & $\mathbf{54.8}$  &  \underline{$56.6$} & $130.24$ &  $238.57$ & $13.45$ & - & - & - & - & - & - & \revised{$65$ h $20$ min} \\ 
ReferFormer+~\cite{ReferFormer} & ResNet-50 & - & HuBERT  & - & - & - & - & - & -  &  \underline{$47.7$}  & \underline{$47.4$}  & \underline{$47.9$} & $138.97$ & $354.81$ & $13.26$ & \revised{$68$ h $36$ min} \\ 
ReferFormer++~\cite{ReferFormer} & ResNet-50 & RoBERTa & HuBERT  & - & - & - & - & - & -  & $40.8$  & $40.2$  & $41.4$ & $224.81$ & $456.67$ & $11.70$ & \revised{$89$ h $21$ min}  \\ 
\rowcolor[gray]{.9}\textbf{EPCFormer} (ours) & ResNet-50  & BERT & - &  $\mathbf{55.6}$ & \underline{$53.9$}  & $\mathbf{57.2}$ & $130.78$ & $273.52$ &  $15.11$ & - &  - & -  & - & - & - & \revised{$72$ h $17$ min}  \\ 
\toprule [1pt]
\rowcolor[gray]{.9} \mygray{\textbf{EPCFormer$^\dagger $} (ours)} & \mygray{ResNet-50} & \mygray{BERT}  &  \mygray{HuBERT}  & \mygray{$55.7$} & \mygray{$54.0$} & \mygray{$57.3$} & \mygray{$229.28$} & \mygray{$488.94$} & \mygray{$12.05$} & \mygray{$59.8$} & \mygray{$58.7$} & \mygray{$60.9$} & \mygray{$229.28$} & \mygray{$488.94$} & \mygray{$12.05$} & \revised{$72$ h $17$ min}  \\ 
\toprule [2pt]
\end{tabular}
}
%</tab-hubert>
\label{tab:hubert}
\end{table*}

\begin{table*} [t!]
\centering
\renewcommand\arraystretch{1.0}
\setlength\tabcolsep{7pt}
\caption{Ablation study of the proposed EVA, EA, and EQ on Ref-Youtube-VOS~\cite{URVOS} and A-Youtube-VOS~\cite{pan2022wnet} datasets.}
\resizebox{\linewidth}{!}{
\begin{tabular}{ccc|cccccc|ccccccccc}
\toprule [2pt]
\multirow{2}{*}{EVA} & \multirow{2}{*}{EA} & \multirow{2}{*}{EQ} & \multicolumn{6}{c|}{Ref-Youtube-VOS} & \multicolumn{6}{c}{A-Youtube-VOS} \\
\cline{4-15}
\specialrule{0em}{1pt}{1pt}
 &&& $\mathcal{J\&F}$ & $\mathcal{J}$ & $\mathcal{F}$ & Param/M & FLOPs/G & FPS & $\mathcal{J\&F}$ & $\mathcal{J}$ & $\mathcal{F}$ & Param/M & FLOPs/G & FPS \\
\toprule [1pt]
        ~ & ~ & ~ & $53.7$  & $52.2$  & $55.1$   & $120.15$ & $267.07$ & $15.37$ & $50.6$  & $49.5$  & $51.6$  & $132.42$ & $388.96$ & $14.05$ \\ 
        \ding{52}  & ~ & ~ & $54.6$  & $53.1$  & $56.1$ & $129.60$  & $273.52$  & $15.28$ & $52.4$  & $51.3$  & $53.4$  & $141.87$ & $395.41$ & $13.97$ \\ 
        ~ & \ding{52} & ~ & $55.2$  & $53.5$  & $56.8$   & $121.34$ & $267.08$ & $15.20$ & $52.0$  & $51.0$  & $53.0$ & $133.60$ & $388.97$ &  $13.88$ \\ 
        ~ & ~ & \ding{52} & $54.8$  & $53.4$  & $56.2$    & $121.34$ & $267.08$ & $15.20$ & $52.2$  & $51.4$  & $53.1$   &  $133.60$ & $388.97$ & $13.88$    \\ 
        \ding{52} & \ding{52} & ~ & $55.0$  & $53.5$  & $56.5$   & $130.78$ & $273.52$ & $15.11$  & $53.1$  & $51.9$  & $54.2$ & $143.05$ & $395.41$ & $13.75$  \\ 
        \ding{52} & ~ & \ding{52} & $54.9$  & $53.3$  & $56.4$   & $130.78$ & $273.52$ & $15.11$  & $52.4$  & $51.3$  & $53.5$ & $143.05$ & $395.41$ & $13.75$   \\ 
        ~ & \ding{52} & \ding{52} & $55.1$  & $53.7$  & $56.6$   & $121.34$ & $267.08$ & $15.20$  & $52.8$  & $51.6$ & $54.0$ & $133.60$ & $388.97$ & $13.88$ \\ 
        \rowcolor[gray]{.9}\ding{52} & \ding{52} & \ding{52} & $\mathbf{55.6}$  & $\mathbf{53.9}$  & $\mathbf{57.2}$   & $130.78$ & $273.52$ & $15.11$ & $\mathbf{53.7}$  & $\mathbf{52.4}$  & $\mathbf{55.0}$ & $143.05$ & $395.41$ & $13.75$   \\ 
\toprule [2pt]
\end{tabular}
}
\label{tab:ablation}
\end{table*}

\subsection{Comparison with State-of-the-Art R-VOS Methods} \label{subsec:sota-rvos}

Tables~\ref{tab:a2d},~\ref{tab:jhmdb}, and~\ref{tab:rvos} list results of different R-VOS methods on A2D-Sentences~\cite{GavrilyukGLS18}, J-HMDB-Sentences~\cite{GavrilyukGLS18}, and Ref-Youtube-VOS~\cite{URVOS}, respectively. 
The results show the performance of EPCFormer is competitive. 
The reason behind this lies in the introduced alignment and well-exploited complementarity between audio and text, which leads to the localization of crucial information in the text.
In summary, the following observations lead to the \revised{following} conclusions:
1) As shown in Table~\ref{tab:a2d}, EPCFormer leads to a higher mAP of $51.7\%$ compared with other methods using CNNs as the backbone on A2D-Sentences.
When compared with the methods using a Transformer as the backbone, EPCFormer surpasses the nearest competitor SgMg~\cite{miao2023spectrum}, by $0.7\%$ in Overall IoU and $0.6\%$ in Mean IoU.
2) In Table~\ref{tab:jhmdb}, with a ResNet-50 backbone~\cite{ResNet}, EPCFormer leads to a higher mAP of $42.8\%$ on J-HMDB-Sentences. 
Compared with the most advanced SgMg~\cite{miao2023spectrum} with the Video-Swin-T~\cite{liu2022video}, EPCFormer with the ViT-Huge backbone~\cite{ViT} brings an improvement of $0.5\%$ in mAP.
3) As shown in Table~\ref{tab:rvos}, EPCFormer using a CNN as the backbone outperforms the previous state-of-the-art method ReferFormer~\cite{ReferFormer} in terms of $\mathcal{F}$ on Ref-Youtube-VOS. 
When using a larger backbone, the performance of EPCFormer further boosts to \revised{an} overall $\mathcal{J\&F}$ of $65.0\%$, outperforming the strong contender VLT ~\cite{vlt} by $1.2\%$.
\revised{Figure}~\ref{fig:rvos} displays some \revised{examples} produced by EPCFormer and ReferFormer~\cite{ReferFormer}.
The results showcase \revised{the} superior language understanding \revised{performance} of the proposed method.
In the second video, EPCFormer precisely localizes and segments the referred rider by accurately focusing on the keyword ``white helmet and clothes" in the sentence, avoiding confusion with other riders.
\revised{EPCFormer also provides accurate predictions for hard-to-discriminate referred objects.}

\subsection{Model Generalizability Analysis}
\label{subsec:analysis}

To investigate the generalizability of different representative existing methods, we established several groups of baselines, detailed as follows:
\textit{1) Wnet+}~\cite{pan2022wnet}: This is the extension of Wnet~\cite{pan2022wnet}, where the MFCC layer~\cite{bouchakour2018mfccs} is replaced with a HuBERT-Base~\cite{hsu2021hubert} as the audio encoder.
This baseline evaluates the impact of various audio decoding methods on A-VOS;
\textit{2) ReferFormer+}~\cite{ReferFormer}: To adapt to A-VOS, this is the extension of~\cite{ReferFormer} where the original text encoder is replaced with a HuBERT-Base~\cite{hsu2021hubert} as the audio encoder.
This tests the generalizability of R-VOS approaches for A-VOS;
\textit{3) Wnet++}~\cite{pan2022wnet}: To adapt to R-VOS, this is the extension of~\cite{pan2022wnet} where the original audio encoder is replaced with a BERT-Base~\cite{BERT} as the text encoder.
It evaluates the generalizability of A-VOS approaches for R-VOS;
\textit{4) ReferFormer++}~\cite{ReferFormer}: This is the augmentation of~\cite{ReferFormer} with an additional HuBERT-Base~\cite{hsu2021hubert} as the ASR preprocessing for raw audio transcriptions.
This baseline evaluates the effectiveness of using ASR as an audio preprocessing step to directly adapt R-VOS models to A-VOS;
\textit{5) EPCFormer+}: The audio encoder originally designed for EPCFormer is substituted with MFCC~\cite{bouchakour2018mfccs}.
This evaluates the impact of various audio encodings on the generalist model.

Table~\ref{tab:hubert} lists the \revised{result} of EPCFormer and \revised{other methods} on A-Youtube-VOS~\cite{pan2022wnet} and Ref-Youtube-VOS~\cite{URVOS}.
We can \revised{make} the following observations from Table~\ref{tab:hubert}. 
First, the proposed EPCFormer maximizes the capabilities of various audio encoders (\textit{e.g.}, MFCC~\cite{bouchakour2018mfccs} or HuBERT~\cite{hsu2021hubert}) to achieve optimal performance on A-VOS.
On one hand, it can be seen that EPCFormer+ with MFCC audio encoder achieves $6.2\%$, $5.4\%$, and $7.1\%$ higher results in terms of $\mathcal{J\&F}$, $\mathcal{J}$, and $\mathcal{F}$ compared with Wnet~\cite{pan2022wnet} on A-Youtube-VOS, respectively.
On the other hand, compared with Wnet+~\cite{pan2022wnet} and ReferFormer+~\cite{ReferFormer}, our EPCFormer achieves at least $10.6\%$ and $5.0\%$ improvements and at most $13.0\%$ and $7.1\%$ on three evaluation metrics, respectively.
Moreover, comparing the two different encoding \revised{methods}, it is evident that while the introduction of HuBERT results in an increase in the number of parameters, the subsequent gains in performance and FPS are deemed acceptable.
For example, substituting HuBERT for \revised{the} MFCC layer in EPCFormer+, brings a performance gain of $3.9\%$ in terms of $\mathcal{J\&F}$ and \revised{maintains} up to $2.9 \times$ run-time speed ($13.75$ FPS \textit{vs.} $4.68$ FPS).

Second, in comparison with the extended models, Table~\ref{tab:hubert} displays that EPCFormer outperforms Wnet++ by $8.8\%$ in $\mathcal{J\&F}$ on R-VOS. In addition, it can be seen that EPCFormer outperforms ReferFormer+ under all the metrics on A-VOS. The reason behind these phenomena lies in the tailored designs for individual tasks, while EPCFormer seamlessly transfers generalized knowledge between A-VOS and R-VOS.

%<*r15-speed>
Third, compared with ReferFormer++~\cite{ReferFormer} using an ASR as audio preprocessing on A-Youtube-VOS, the proposed EPCFormer is $12.9\%$ higher than it at the overall $\mathcal{J\&F}$ while reducing both the FLOPs cost and parameters by $61.26$G and $81.76$M, respectively.
\revised{This indicates that in comparison to the paradigm of first transcribing with ASR and then segmenting based on text, the proposed EPCFormer reflects an advantage in both accuracy and speed.}
%</r15-speed> 
This phenomenon \revised{can} be attributed to the non-end-to-end training of the former paradigm, potentially leading to suboptimal solutions.

Fourth, it can be observed that \revised{the} simultaneous input of two modalities into EPCFormer further enhances accuracy.
This indicates the generalist ability of EPCFormer, which is effective not only for text-only and audio-only inputs but also for scenarios involving simultaneous input of two modalities.

\revised{In summary, current R-VOS, A-VOS, and ASR-based methods show ineffectiveness in bridging the gaps between distinct modality tasks.} 
Thanks to the proposed EVA and EA, EPCFormer seamlessly switches between two tasks with guaranteed precision and low computational expense.

\subsection{Ablation Studies}
\label{subsec:ablation}

Tables~\ref{tab:fusion}-\ref{tab:train} present the ablation results of the proposed components, 
including EVA, EA, EQ, and multi-task training \revised{methods}, 
\revised{whereas Table~\ref{tab:backbone} presents the effects of different backbones on our model’s performance.}
Table~\ref{tab:mlp} and Table~\ref{tab:contrast} provide the hyper-parameter analyses.
The results demonstrate the effectiveness of the proposed components, as the overall performance is superior when all components are integrated.

\textbf{Evaluation of EVA.}
\revised{Firstly, comparing the $1^{\mathrm{st}}$ and $2^{\mathrm{rd}}$ rows} of Table~\ref{tab:ablation}, it can be seen that adding EVA only costs $9.45$M parameters but yields 
noteworthy gain of $0.9\%$ and $1.8\%$ in terms of $\mathcal{J\&F}$ on Ref-Youtube-VOS and A-Youtube-VOS, respectively.
This phenomenon shows that EVA effectively fosters interactions among various modality features.
Secondly, from the last and penultimate row, it becomes apparent that, following the integration of EA and EQ, adding EVA contributes to further performance gains.
This implies that EVA's capacity to capture homogeneous semantic information from distinct referring modalities is optimal, particularly in the context of aligned text and audio features.
%<*r13-fusion-2>
\revised{Thirdly, Table~\ref{tab:fusion} shows that using \revised{the} addition operation for the interaction of ATC yields the best performance}.
%</r13-fusion-2>
In addition, Table~\ref{tab:attention} shows the impact of interactions between audio and text modalities on the ATC module in the EVA, where optimal performance is achieved when both audio and text implement bidirectional interaction.

\textbf{Evaluation of EA.}
\revised{Initially, comparing the $1^{\mathrm{st}}$ and $2^{\mathrm{rd}}$ rows} of Table~\ref{tab:ablation} reveals that only adding EA brings a performance gain of $1.5\%$ and $1.4\%$ in terms of $\mathcal{J\&F}$ on Ref-Youtube-VOS and A-Youtube-VOS, respectively, with minimal cost on parameters and negligible impact on FPS.
This implies that adding EA is beneficial for the model to learn homogeneous semantic information in two different modalities.
Secondly, when comparing the \revised{$1^{\mathrm{st}}$ and $4^{\mathrm{th}}$} line, it can be seen that adding EQ increases the $\mathcal{J\&F}$ metric by $1.1\%$ and $1.6\%$ on Ref-Youtube-VOS and A-Youtube-VOS, respectively.
Meanwhile, the results in the \revised{$7^{\mathrm{th}}$ and $3^{\mathrm{rd}}$} rows reveal that the performance is further enhanced by incorporating EA in addition to adding EQ.
The reason behind this phenomenon lies in the benefit derived from using aligned text and audio embeddings for initializing queries.
Finally, a comparison between the \revised{$2^{\mathrm{rd}}$ and $4^{\mathrm{th}}$} rows indicates that incorporating EA on top of adding EVA consistently improves performance on both tasks.
This implies that adding EA is consistently beneficial for the EVA module to learn from different modalities, enhancing its capacity to handle referring text and audio prompts in a unified manner.

\begin{table} [t!]
\centering
\renewcommand\arraystretch{1.0}
\setlength\tabcolsep{7pt}
%<*r13-fusion>
\caption{\revised{Ablation study on different fusion ways for attention  matrix of ATC in the proposed EVA module.}}
\resizebox{\linewidth}{!}{
\begin{tabular}{l|ccc|ccc}
\toprule [2pt]
            \multirow{3}{*}{Fusion Way} & \multicolumn{3}{c|}{A2D-Sentence} & \multicolumn{3}{c}{A-A2D}   \\ 
\cline{2-7}
\specialrule{0em}{1pt}{1pt}
& \multicolumn{2}{c}{IoU} &  \multirow{2}{*}{mAP} & \multirow{2}{*}{$\mathcal{J\&F}$} & \multirow{2}{*}{$\mathcal{J}$} & \multirow{2}{*}{$\mathcal{F}$} \\
\cline{2-3}
& Overall & Mean & & \\
\toprule [1pt]
        Concatenation & $ 73.2$  & $67.5$  & $49.8$  & $60.9$  & $58.9$  & $62.8$   \\ 
        \rowcolor[gray]{.9}
        Addition & $\textbf{74.6}$  & $\textbf{67.9}$  & $\textbf{51.7}$  & $\textbf{63.0}$  & $\textbf{60.7}$  & $\textbf{65.2}$ \\ 
\toprule [2pt]
\end{tabular}
}
%</r13-fusion>
\label{tab:fusion}
\end{table} 

\begin{table} [t!]
\centering
\renewcommand\arraystretch{1.0}
\setlength\tabcolsep{7pt}
\caption{Ablation study on different interactions of ATC in the proposed EVA module. $\nleftrightarrow$ means the attention matrix of ATC is not shared. $\leftarrow$and$\rightarrow$ means the attention matrix of ATC is one-way shared. $\leftrightarrow$ means the attention matrix of ATC is bi-way shared.}
\resizebox{\linewidth}{!}{
\begin{tabular}{c|ccc|ccc}
\toprule [2pt]
            \multirow{2}{*}{Method} & \multicolumn{3}{c|}{Ref-Youtube-VOS } & \multicolumn{3}{c}{A-Youtube-VOS}   \\ 
\cline{2-7}
\specialrule{0em}{1pt}{1pt}
& $\mathcal{J\&F}$ & $\mathcal{J}$ & $\mathcal{F}$& $\mathcal{J\&F}$ & $\mathcal{J}$ & $\mathcal{F}$\\
\toprule [1pt]
        Text$\nleftrightarrow$Audio & $54.1$  & $52.8$  & $55.5$  & $51.2$  & $50.2$  & $52.2$   \\ 
        Text$\leftarrow$Audio & $54.7$  & $53.1$  & $56.2$  & $51.3$  & $50.2$  & $52.4$   \\ 
        Text$\rightarrow$Audio & $54.3$  & $52.8$  & $55.9$  & $53.0$  & $51.8$  & $54.3$   \\ 
        \rowcolor[gray]{.9}Text$\leftrightarrow$Audio & $\mathbf{55.6}$  & $\mathbf{53.9}$  & $\mathbf{57.2}$  & $\mathbf{53.7}$  & $\mathbf{52.4}$  & $\mathbf{55.0}$   \\ 
\toprule [2pt]
\end{tabular}
}
\label{tab:attention}
\end{table} 

\begin{table} [t!]
\centering
\renewcommand\arraystretch{1.0}
\setlength\tabcolsep{7pt}
\caption{Ablation study on different query initialization strategies of EQ on EPCFormer when training.}
\resizebox{\linewidth}{!}{
\begin{tabular}{cc|ccc|ccc}
\toprule [2pt]
\multirow{2}{*}{Audio} & \multirow{2}{*}{Text} & \multicolumn{3}{c|}{Ref-Youtube-VOS} & \multicolumn{3}{c}{A-Youtube-VOS}   \\ 
\cline{3-8}
\specialrule{0em}{1pt}{1pt}
 & & $\mathcal{J\&F}$ & $\mathcal{J}$ & $\mathcal{F}$& $\mathcal{J\&F}$ & $\mathcal{J}$ & $\mathcal{F}$\\
\toprule [1pt]
&&  $55.0$  & $53.5$  & $56.5$ & $53.1$  & $51.9$  & $54.2$   \\
\ding{52} &   & $55.5$  & $53.6$  & $\mathbf{57.3}$ & $53.0$  & $51.8$  & $54.1$   \\ 
& \ding{52} & $54.9$  & $53.3$  & $56.4$  & $53.6$  & $52.4$  & $54.7$   \\ 
\rowcolor[gray]{.9} \ding{52}  & \ding{52} & $\mathbf{55.6}$  & $\mathbf{53.9}$  & $57.2$  & $\mathbf{53.7}$  & $\mathbf{52.4}$  & $\mathbf{55.0}$   \\ 
\toprule [2pt]
\end{tabular}
}

\label{tab:query}
\end{table}

\begin{table} [t!]
\centering
\renewcommand\arraystretch{1.0}
\setlength\tabcolsep{5.5pt}
\caption{Ablation study on different multi-task training settings on EPCFormer. Note that ``Audio'', ``Text'',  and ``Mix'' represent an input that is audio-only, text-only, or a combination of text and audio when training.}
\resizebox{\linewidth}{!}{
\begin{tabular}{ccc|ccc|ccc}
\toprule [2pt]
\multirow{2}{*}{Audio} & \multirow{2}{*}{Text} & \multirow{2}{*}{Mix} & \multicolumn{3}{c|}{Ref-Youtube-VOS} & \multicolumn{3}{c}{A-Youtube-VOS} \\
\cline{4-9}
\specialrule{0em}{1pt}{1pt}
 &&& $\mathcal{J\&F}$ & $\mathcal{J}$ & $\mathcal{F}$& $\mathcal{J\&F}$ & $\mathcal{J}$ & $\mathcal{F}$\\
\toprule [1pt]
\ding{52} &      &      & -  & -  & -  & $50.7$  & $49.5$  & $51.9$  \\ 
 & \ding{52} &  & $53.7$  & $52.1$  & $55.2$  & -  & -  & -  \\ 
\ding{52} & \ding{52}     &  & $53.7$  & $52.3$  & $55.1$  & $51.1$  & $50.1$  & $52.1$   \\ 
\rowcolor[gray]{.9} \ding{52} & \ding{52}     & \ding{52} & $\mathbf{55.6}$  & $\mathbf{53.9}$  & $\mathbf{57.2}$ & $\mathbf{53.7}$  & $\mathbf{52.4}$  & $\mathbf{55.0}$   \\
\toprule [2pt]
\end{tabular}
}
\label{tab:train}
\end{table}

\begin{table} [t!]
\centering
\renewcommand\arraystretch{1.1}
\setlength\tabcolsep{4pt}
%<*tab-backbone>
\caption{\revised{Ablation study on different backbones.}}
\resizebox{\linewidth}{!}{
\begin{tabular}{c|c|ccc|ccc}
\toprule [2pt]
            \multirow{3}{*}{\shortstack{Visual\\Backbone}}  & \multirow{3}{*}{\shortstack{Text\\Backbone}} & \multicolumn{3}{c|}{A2D-Sentence} & \multicolumn{3}{c}{A-A2D}   \\ 
\cline{3-8}
\specialrule{0em}{1pt}{1pt}
&& \multicolumn{2}{c}{IoU} &  \multirow{2}{*}{mAP} & \multirow{2}{*}{$\mathcal{J\&F}$} & \multirow{2}{*}{$\mathcal{J}$} & \multirow{2}{*}{$\mathcal{F}$} \\
\cline{3-4}
&& Overall & Mean & & \\
\toprule [1pt]
       Video-Swin-B & RoBERTa & $80.3$  & $72.2$  & $58.0$  & $63.6$  & $61.4$  & $65.7$ \\ 
       Video-Swin-B & BERT & $80.5$  & $72.0$  & $58.2$  & $63.2$  & $61.1$  & $65.3$ \\ 
       ViT-H & RoBERTa & $80.1$  & $\textbf{72.7}$  & $58.1$  & $64.5$ & $62.0$ &$66.9$  \\ 
       \rowcolor[gray]{.9} ViT-H & BERT & $\textbf{80.6}$  & $72.6$  & $\textbf{58.2}$  & $\textbf{64.9}$  & $\textbf{62.6}$  & $\textbf{67.3}$ \\ 
       
\toprule [2pt]
\end{tabular}
}
%</tab-backbone>
\label{tab:backbone}
\end{table} 

\begin{table} [t!]
\centering
\renewcommand\arraystretch{1.0}
\setlength\tabcolsep{9pt}
\caption{Hyper-parameter analysis on the numbers of MLP layers for ECL in the proposed EA module.}
\resizebox{\linewidth}{!}{
\begin{tabular}{c|ccc|ccc}
\toprule [2pt]
            \multirow{2}{*}{Layer} & \multicolumn{3}{c|}{Ref-Youtube-VOS} & \multicolumn{3}{c}{A-Youtube-VOS}   \\ 
\cline{2-7}
\specialrule{0em}{1pt}{1pt}
& $\mathcal{J\&F}$ & $\mathcal{J}$ & $\mathcal{F}$& $\mathcal{J\&F}$ & $\mathcal{J}$ & $\mathcal{F}$\\
\toprule [1pt]
$1$&        $54.6$  & $53.2$  & $56.1$  & $53.4$  & $52.3$  & $54.5$   \\ 
\rowcolor[gray]{.9}2&         $\mathbf{55.6}$  & $\mathbf{53.9}$  & $\mathbf{57.2}$  & $\mathbf{53.7}$  & $\mathbf{52.4}$  & $\mathbf{55.0}$   \\
$3$&       $54.7$  & $53.1$  & $56.2$   & $53.2$  & $52.1$  & $54.3$   \\ 
$4$&       $55.2$  & $53.7$  & $56.8$   & $53.6$  & $52.4$  & $54.7$   \\ 
\toprule [2pt]
\end{tabular}
}
\label{tab:mlp}
\end{table}

\begin{table} [t!]
\centering
\renewcommand\arraystretch{1.0}
\setlength\tabcolsep{9pt}
\caption{Hyper-parameter analysis on the loss weight $\lambda_{expr}$.}
\resizebox{\linewidth}{!}{
\begin{tabular}{c|ccc|ccc}
\toprule [2pt]
            \multirow{2}{*}{$\lambda_{expr}$} & \multicolumn{3}{c|}{Ref-Youtube-VOS } & \multicolumn{3}{c}{A-Youtube-VOS}   \\ 
\cline{2-7}
\specialrule{0em}{1pt}{1pt}
& $\mathcal{J\&F}$ & $\mathcal{J}$ & $\mathcal{F}$& $\mathcal{J\&F}$ & $\mathcal{J}$ & $\mathcal{F}$\\
\toprule [1pt]
        $0$ & $54.7$  & $53.2$  & $56.2$  & $52.4$  & $51.3$  & $53.5$   \\ 
        $0.5$ & $54.6$  & $53.1$  & $56.1$  & $53.5$  & $\mathbf{52.5}$  & $54.4$   \\ 
        \rowcolor[gray]{.9}1 & $\mathbf{55.6}$ & $\mathbf{53.9}$  & $\mathbf{57.2}$  & $\mathbf{53.7}$  & $52.4$  & $\mathbf{55.0}$   \\ 
        $1.5$ & $55.0$  & $53.4$  & $56.6$  & $53.1$  & $52.1$  & $54.2$   \\ 
\toprule [2pt]
\end{tabular}
}
\label{tab:contrast}
\end{table}

\textbf{Evaluation of distinct strategies of EQ.}
As evidenced in Table~\ref{tab:query}, 
three distinct strategies are employed to assist in initializing the segmentation Transformer decoder's input queries.
From top to bottom are without EQ, using only text embeddings $\mathcal{E}_t$, using only audio embeddings $\mathcal{E}_a$, or using both $\mathcal{E}_t$ and $\mathcal{E}_a$ as queries, respectively.
The results showcase the positive impact of employing the EQ strategies both on the audio and text modalities, where the best results are obtained in both A-VOS and R-VOS.
In contrast to the other scenarios (from first to third rows of Table~\ref{tab:query}), improvements of $0.6\%$, $0.1\%$, and $0.7\%$ are observed in terms of $\mathcal{J\&F}$ on Ref-Youtube-VOS, and $0.6\%$, $0.7\%$, and $0.1\%$ on A-Youtube-VOS, respectively.

\textbf{Evaluation of multi-task training.}
In Table~\ref{tab:train}, four training strategies are established to evaluate the impact of multi-task training.
From top to bottom are audio-only training, text-only training, audio-only and text-only training, and the proposed multi-task training (text-only, audio-only, and a combination of both text and audio training).
The result of Table~\ref{tab:train} reveals that the proposed multi-task training method maximizes the potential of the proposed EPCFomer, 
yielding optimal results compared to other training methods on both A-VOS and R-VOS tasks.
This conclusion can be drawn from the following observations: 
1) In the \revised{$1^{\mathrm{st}}$ and $3^{\mathrm{th}}$} rows, the result on every metric for the audio-only and text-only training is on par with the text-only training on Ref-Youtube-VOS. 
Notably, a marginal increase of only $0.4\%$ in $\mathcal{J\&F}$ is attained when compared with the audio-only training on A-Youtube-VOS.
2) Comparing the last two rows, it can be observed that the proposed multi-task training yields results over $1.9\%$ and $2.6\%$ higher in terms of $\mathcal{J\&F}$ on Ref-Youtube-VOS and A-Youtube-VOS, respectively.

%<*r24-backbone>
\revised{\textbf{Evaluation of distinct backbones.}
Table~\ref{tab:backbone} shows the effect of different visual encoders and text encoders, namely Video-Swin-B~\cite{liu2022video}, ViT-Huge~\cite{ViT}, RoBERTa~\cite{roberta} and BERT~\cite{BERT}.
It can be observed that our model exhibits competitive results across different backbones, leading to the best result with ViT-Huge and BERT. This is because this combination excels in multi-modal interaction scenarios, particularly with the inclusion of audio references.}
%</r24-backbone>

\textbf{Hyper-parameter analysis.}
Tables~\ref{tab:mlp} and~\ref{tab:contrast} list the results of different MLP layers and the weight $\lambda_{expr}$ in contrastive loss on the proposed EPCFormer, respectively.
The results reveal that the model performs optimally with $2$ MLP layers and $\lambda_{expr}=1$.
As illustrated in Table~\ref{tab:mlp}, it is evident that augmenting the number of MLP layers from $1$ to $2$ results in enhanced performance in both tasks.
However, when the number of layers reaches $3$, the performance decline is observed. 
Upon comparing the model's performance for varying $\lambda_{expr}$ weight values, \textit{i.e.}, $\lambda_{expr}=0,0.5,1,1.5$, as shown in Table~\ref{tab:contrast}, it becomes apparent that $\lambda_{expr}=1$ yields the most favorable outcomes.

\begin{figure}[t!]
\centering
\includegraphics[width=1.0\linewidth]{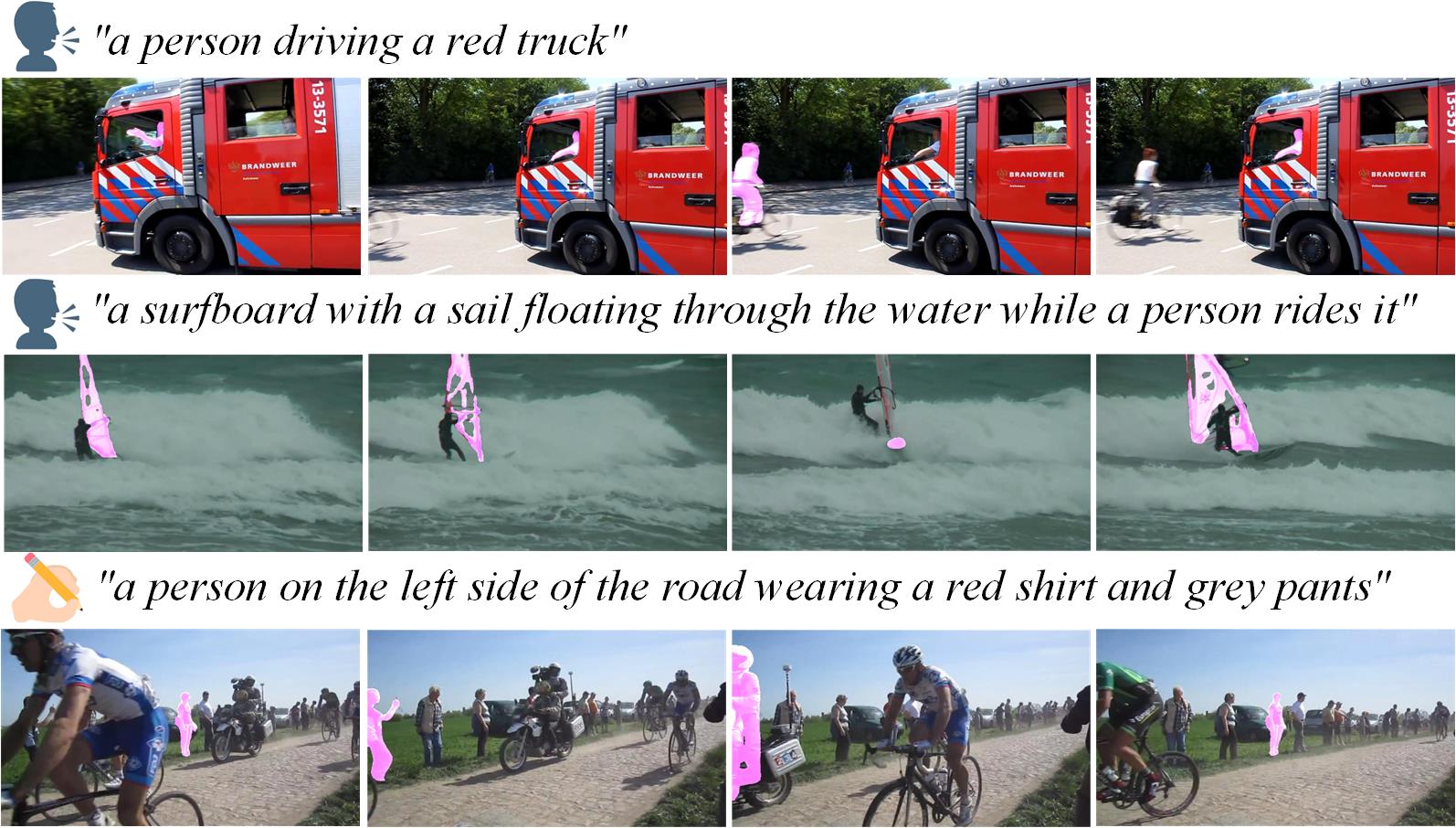}
%<*fig-limit>
\caption{\revised{Visualization of failure cases of the proposed EPCFomrer on A-Youtube-VOS~\cite{pan2022wnet} and Ref-Youtube-VOS~\cite{URVOS}.}}
%</fig-limit>
\label{fig:failure_case}
\end{figure}

\subsection{\revised{Limitation Analysis}} \label{subsec:limitation}
%<*r16-limit>
\revised{We examine the typical failure instances.
For instance, in $1^{\mathrm{st}}$ example of Figure~\ref{fig:failure_case}, the audio asks us to locate the person whose lower body is obstructed by the truck.
EPCFormer experiences misdetection for the passing rider.
In $2^{\mathrm{nd}}$ example, the atypical slender sail causes the model to produce an omission.
In $3^{\mathrm{rd}}$ instance where numerous persons are present, EPCFormer encounters challenges in localizing the referred occluded person due to insufficient temporal context analysis.}
%</r16-limit>

\section{Conclusions}
\label{sec:conclusion}

\revised{In this paper, we propose EPCFormer, a novel framework for universal referring video object segmentation. Our method effectively analyzes audio and text to guide the segmentation of referred objects.} Firstly, we introduce the EA based on contrastive learning to align semantically related audio and text. This module supervises the linear mapping of both modalities into a multi-modal embedding space. \revised{Secondly, to achieve comprehensive interaction among audio, text, and visual modalities, we propose the EVA module to further explore the complementary relationship and generate a precise mask.} Extensive experiments demonstrate that EPCFormer achieves state-of-the-art performance on R-VOS and A-VOS tasks.

\revised{Moreover, the primary application scenario for our EPCFormer involves human-computer interaction. 
Dealing with above situations and developing a lightweight network tailored for edge devices could be interesting topics for future works.}
%These leave space for further investigation in our future work.}

\section*{Acknowledgment}
\label{sec:findings}

This work was supported in part by the National Natural Science Foundation of China under Grant U21A20518, Grant 62106071, Grant U23A20341, and Grant 62473139; and in part by Hangzhou SurImage Technology Company Ltd.

%% If you have bibdatabase file and want bibtex to generate the
%% bibitems, please use
%%
%%  \bibliographystyle{elsarticle-harv} 
%%  \bibliography{<your bibdatabase>}

%% else use the following coding to input the bibitems directly in the
%% TeX file.

%\begin{thebibliography}{00}

%% \bibitem[Author(year)]{label}
%% Text of bibliographic item

%\bibitem[ ()]{}
\bibliographystyle{elsarticle-num}  
% \bibliographystyle{unsrt}
%这里不用改，对应的是elsarticle-num.bst文件
\bibliography{mybibfile}  

\begin{thebibliography}{10}
\expandafter\ifx\csname url\endcsname\relax
  \def\url#1{\texttt{#1}}\fi
\expandafter\ifx\csname urlprefix\endcsname\relax\def\urlprefix{URL }\fi
\expandafter\ifx\csname href\endcsname\relax
  \def\href#1#2{#2} \def\path#1{#1}\fi

\bibitem{pan2022wnet}
W.~Pan, et~al., Wnet: Audio-guided video object segmentation via wavelet-based cross-modal denoising networks, in: Proc. IEEE Conf. Comput. Vis. Pattern Recognit. (CVPR), 2022, pp. 1310--1321.

\bibitem{GavrilyukGLS18}
K.~Gavrilyuk, A.~Ghodrati, Z.~Li, C.~G.~M. Snoek, Actor and action video segmentation from a sentence, in: Proc. IEEE Conf. Comput. Vis. Pattern Recognit. (CVPR), 2018, pp. 5958--5966.

\bibitem{MTTR}
A.~Botach, E.~Zheltonozhskii, C.~Baskin, End-to-end referring video object segmentation with multimodal transformers, in: Proc. IEEE Conf. Comput. Vis. Pattern Recognit. (CVPR), 2022, pp. 4975--4985.

\bibitem{ReferFormer}
J.~Wu, Y.~Jiang, P.~Sun, Z.~Yuan, P.~Luo, Language as queries for referring video object segmentation, in: Proc. IEEE Conf. Comput. Vis. Pattern Recognit. (CVPR), 2022, pp. 4964--4974.

\bibitem{zhang2024video}
Y.~Zhang, Z.~Zhang, M.~Liao, S.~Tian, R.~You, W.~Zou, C.~Xu, Video generalized semantic segmentation via non-salient feature reasoning and consistency, Knowl-Based Syst. (2024) 111584.

\bibitem{nake2001human}
F.~Nake, S.~Grabowski, Human--computer interaction viewed as pseudo-communication, Knowl-Based Syst. 14~(8) (2001) 441--447.

\bibitem{xie2024satr}
J.~Xie, J.~Liu, G.~Wang, F.~Zhou, Satr: Semantics-aware triadic refinement network for referring image segmentation, Knowl-Based Syst. 284 (2024) 111243.

\bibitem{wang2019acan}
H.~Wang, C.~Deng, J.~Yan, D.~Tao, Asymmetric cross-guided attention network for actor and action video segmentation from natural language query, in: Proc. IEEE Int. Conf. Comput. Vis. (ICCV), 2019, pp. 3938--3947.

\bibitem{khoreva2018video}
A.~Khoreva, A.~Rohrbach, B.~Schiele, Video object segmentation with language referring expressions, in: Proc. Asi. Conf. Comput. Vis. (ACCV), Vol. 11364, 2018, pp. 123--141.

\bibitem{jing2021locate}
Y.~Jing, T.~Kong, W.~Wang, L.~Wang, L.~Li, T.~Tan, Locate then segment: A strong pipeline for referring image segmentation, in: Proc. IEEE Conf. Comput. Vis. Pattern Recognit. (CVPR), 2021, pp. 9858--9867.

\bibitem{LIN2023120960}
J.~Lin, et~al., {BRPPNet:} {Balanced} privacy protection network for referring personal image privacy protection, Expert Syst. Appl. (2023) 120960.

\bibitem{schneider2019wav2vec}
S.~Schneider, A.~Baevski, R.~Collobert, M.~Auli, wav2vec: Unsupervised pre-training for speech recognition, Proc. Annu. Conf. Int. Speech Commun. Assoc. (2019) 3465--3469.

\bibitem{hsu2021hubert}
W.-N. Hsu, B.~Bolte, Y.-H.~H. Tsai, K.~Lakhotia, R.~Salakhutdinov, A.~Mohamed, {HuBERT:} {Self-supervised} speech representation learning by masked prediction of hidden units, IEEE/ACM Trans. Audio Speech Lang. Process. 29 (2021) 3451--3460.

\bibitem{ning2022audio}
H.~Ning, B.~Zhao, Z.~Hu, L.~He, E.~Pei, Audio--visual collaborative representation learning for dynamic saliency prediction, Knowl-Based Syst. 256 (2022) 109675.

\bibitem{zhou2022audio}
J.~Zhou, et~al., Audio-visual segmentation, in: Proc. Eur. Conf. Comput. Vis. (ECCV), Vol. 13697, 2022, pp. 386--403.

\bibitem{gao2023avsegformer}
S.~Gao, Z.~Chen, G.~Chen, W.~Wang, T.~Lu, {AVSegFormer:} {Audio-visual} segmentation with transformer, arXiv preprint arXiv:2307.01146 (2023).

\bibitem{URVOS}
S.~Seo, J.-Y. Lee, B.~Han, {URVOS:} {Unified} referring video object segmentation network with a large-scale benchmark, in: Proc. Eur. Conf. Comput. Vis. (ECCV), Vol. 12360, 2020, pp. 208--223.

\bibitem{liu2022cmpc}
S.~Liu, T.~Hui, S.~Huang, Y.~Wei, B.~Li, G.~Li, Cross-modal progressive comprehension for referring segmentation, IEEE Trans. Pattern Anal. Mach. Intell. (2022) 4761--4775.

\bibitem{miao2023spectrum}
B.~Miao, M.~Bennamoun, Y.~Gao, A.~Mian, Spectrum-guided multi-granularity referring video object segmentation, in: Proc. IEEE Int. Conf. Comput. Vis. (ICCV), 2023, pp. 920--930.

\bibitem{hui2021collaborative}
T.~Hui, et~al., Collaborative spatial-temporal modeling for language-queried video actor segmentation, in: Proc. IEEE Conf. Comput. Vis. Pattern Recognit. (CVPR), 2021, pp. 4187--4196.

\bibitem{liang2021clawcranenet}
C.~Liang, Y.~Wu, Y.~Luo, Y.~Yang, {ClawCraneNet:} {Leveraging} object-level relation for text-based video segmentation, arXiv preprint arXiv:2103.10702 (2021).

\bibitem{YOFO}
D.~Li, et~al., You only infer once: Cross-modal meta-transfer for referring video object segmentation, in: Proc. Conf. Artif. Intell. (AAAI), Vol.~36, 2022, pp. 1297--1305.

\bibitem{wu2022multi}
D.~Wu, X.~Dong, L.~Shao, J.~Shen, Multi-level representation learning with semantic alignment for referring video object segmentation, in: Proc. IEEE Conf. Comput. Vis. Pattern Recognit. (CVPR), 2022, pp. 4986--4995.

\bibitem{vaswani2017attention}
A.~Vaswani, et~al., Attention is all you need, in: Proc. Adv. Neural Inform. Process. Syst., Vol.~30, 2017, pp. 5998--6008.

\bibitem{prpe}
K.~Ning, L.~Xie, F.~Wu, Q.~Tian, Polar relative positional encoding for video-language segmentation, in: Proc. Int. Joint Conf. Artif. Intell. (IJCAI), 2020, pp. 948--954.

\bibitem{ye2021referring}
L.~Ye, M.~Rochan, Z.~Liu, X.~Zhang, Y.~Wang, Referring segmentation in images and videos with cross-modal self-attention network, IEEE Trans. Pattern Anal. Mach. Intell. 44~(7) (2022) 3719--3732.

\bibitem{LBDT}
Z.~Ding, T.~Hui, J.~Huang, X.~Wei, J.~Han, S.~Liu, Language-bridged spatial-temporal interaction for referring video object segmentation, in: Proc. IEEE Conf. Comput. Vis. Pattern Recognit. (CVPR), 2022, pp. 4954--4963.

\bibitem{vlt}
H.~Ding, C.~Liu, S.~Wang, X.~Jiang, {VLT:} {Vision-language} transformer and query generation for referring segmentation, IEEE Trans. Pattern Anal. Mach. Intell. 45~(6) (2023) 7900--7916.

\bibitem{DETR}
N.~Carion, F.~Massa, G.~Synnaeve, N.~Usunier, A.~Kirillov, S.~Zagoruyko, End-to-end object detection with transformers, in: Proc. Eur. Conf. Comput. Vis. (ECCV), Vol. 12346, 2020, pp. 213--229.

\bibitem{DeformableDETR}
X.~Zhu, W.~Su, L.~Lu, B.~Li, X.~Wang, J.~Dai, Deformable {DETR}: {Deformable} transformers for end-to-end object detection, in: Proc. Int. Conf. Learn. Represent. (ICLR), 2021.

\bibitem{VISTR}
Y.~Wang, et~al., End-to-end video instance segmentation with transformers, in: Proc. IEEE Conf. Comput. Vis. Pattern Recognit. (CVPR), 2021, pp. 8741--8750.

\bibitem{wu2023online}
D.~Wu, T.~Wang, Y.~Zhang, X.~Zhang, J.~Shen, {OnlineRefer:} {A} simple online baseline for referring video object segmentation, in: Proc. IEEE Int. Conf. Comput. Vis. (ICCV), 2023, pp. 2761--2770.

\bibitem{Mask2Former}
B.~Cheng, I.~Misra, A.~G. Schwing, A.~Kirillov, R.~Girdhar, Masked-attention mask transformer for universal image segmentation, in: Proc. IEEE Conf. Comput. Vis. Pattern Recognit. (CVPR), 2022, pp. 1280--1289.

\bibitem{yan2023referred}
S.~Yan, et~al., Referred by multi-modality: A unified temporal transformer for video object segmentation, arXiv preprint arXiv:2305.16318 (2023).

\bibitem{hadsell2006dimensionality}
R.~Hadsell, S.~Chopra, Y.~LeCun, Dimensionality reduction by learning an invariant mapping, in: Proc. IEEE Conf. Comput. Vis. Pattern Recognit. (CVPR), Vol.~2, 2006, pp. 1735--1742.

\bibitem{CLIP}
A.~Radford, et~al., Learning transferable visual models from natural language supervision, in: Proc. Int. Conf. Mach. Learn. (ICML), Vol. 139, 2021, pp. 8748--8763.

\bibitem{ying2023ctvis}
K.~Ying, et~al., {CTVIS:} {Consistent} training for online video instance segmentation, in: Proc. IEEE Int. Conf. Comput. Vis. (ICCV), 2023, pp. 899--908.

\bibitem{zhong2023contrast}
G.~Zhong, J.~Yuan, P.~Wang, K.~Yang, W.~Guan, Z.~Li, Contrast-augmented diffusion model with fine-grained sequence alignment for markup-to-image generation, in: Proc. ACM Int. Conf. Multimedia, 2023, pp. 5311--5320.

\bibitem{IDOL}
J.~Wu, Q.~Liu, Y.~Jiang, S.~Bai, A.~Yuille, X.~Bai, In defense of online models for video instance segmentation, in: Proc. Eur. Conf. Comput. Vis. (ECCV), Vol. 13688, 2022, pp. 588--605.

\bibitem{cris}
Z.~Wang, et~al., {CRIS:} {CLIP-driven} referring image segmentation, in: Proc. IEEE Conf. Comput. Vis. Pattern Recognit. (CVPR), 2022, pp. 11676--11685.

\bibitem{luo2023soc}
Z.~Luo, et~al., {SOC}: {Semantic-assisted} object cluster for referring video object segmentation, in: Proc. Adv. Neural Inform. Process. Syst., 2023.

\bibitem{CLAP}
Y.~Wu, K.~Chen, T.~Zhang, Y.~Hui, T.~Berg-Kirkpatrick, S.~Dubnov, Large-scale contrastive language-audio pretraining with feature fusion and keyword-to-caption augmentation, in: Proc. IEEE Int. Conf. Acoust., Speech Signal Process. (ICASSP), 2023, pp. 1--5.

\bibitem{bapna2022mslam}
A.~Bapna, et~al., {mSLAM:} {Massively} multilingual joint pre-training for speech and text, arXiv preprint arXiv:2202.01374 (2022).

\bibitem{zhang2022speechlm}
Z.~Zhang, et~al., {SpeechLM:} {Enhanced} speech pre-training with unpaired textual data, arXiv preprint arXiv:2209.15329 (2022).

\bibitem{jia2021scaling}
C.~Jia, et~al., Scaling up visual and vision-language representation learning with noisy text supervision, in: Proc. Int. Conf. Mach. Learn. (ICML), 2021, pp. 4904--4916.

\bibitem{zhu2023vatlm}
Q.~Zhu, et~al., {VATLM:} {Visual-audio-text} pre-training with unified masked prediction for speech representation learning, IEEE Trans. Multimedia (2023).

\bibitem{guzhov2022audioclip}
A.~Guzhov, F.~Raue, J.~Hees, A.~Dengel, Audioclip: Extending clip to image, text and audio, in: Proc. IEEE Int. Conf. Acoust., Speech Signal Process. (ICASSP), 2022, pp. 976--980.

\bibitem{SpeechCLIP}
Y.-J. Shih, H.-F. Wang, H.-J. Chang, L.~Berry, H.-y. Lee, D.~Harwath, {SpeechCLIP:} {Integrating} speech with pre-trained vision and language model, in: Proc. IEEE Spoken Lang. Technol. Workshop, 2023, pp. 715--722.

\bibitem{zhang2023cmx}
J.~Zhang, H.~Liu, K.~Yang, X.~Hu, R.~Liu, R.~Stiefelhagen, {CMX:} {Cross-modal} fusion for {RGB-X} semantic segmentation with transformers, IEEE Trans. Intell. Transp. Syst. 24~(12) (2023) 14679--14694.

\bibitem{zhang2023delivering}
J.~Zhang, et~al., Delivering arbitrary-modal semantic segmentation, in: Proc. IEEE Conf. Comput. Vis. Pattern Recognit. (CVPR), 2023, pp. 1136--1147.

\bibitem{gu2023dataseg}
X.~Gu, et~al., {DaTaSeg:} {Taming} a universal multi-dataset multi-task segmentation model, arXiv preprint arXiv:2306.01736 (2023).

\bibitem{zhang2023vpuformer}
X.~Zhang, K.~Yang, J.~Lin, J.~Yuan, Z.~Li, S.~Li, {VPUFormer:} {Visual} prompt unified transformer for interactive image segmentation, arXiv preprint arXiv:2306.06656 (2023).

\bibitem{wang2023seggpt}
X.~Wang, X.~Zhang, Y.~Cao, W.~Wang, C.~Shen, T.~Huang, {SegGPT:} {Segmenting} everything in context, in: Proc. IEEE Int. Conf. Comput. Vis. (ICCV), 2023.

\bibitem{wu2023segment_every_reference}
J.~Wu, Y.~Jiang, B.~Yan, H.~Lu, Z.~Yuan, P.~Luo, Segment every reference object in spatial and temporal spaces, in: Proc. IEEE Int. Conf. Comput. Vis. (ICCV), 2023, pp. 2538--2550.

\bibitem{GLIP}
L.~Li, et~al., Grounded language-image pre-training, in: Proc. IEEE Conf. Comput. Vis. Pattern Recognit. (CVPR), 2022, pp. 10955--10965.

\bibitem{zhang2021k}
W.~Zhang, J.~Pang, K.~Chen, C.~C. Loy, {K-Net:} {Towards} unified image segmentation, in: Proc. Adv. Neural Inform. Process. Syst., Vol.~34, 2021, pp. 10326--10338.

\bibitem{cheng2021maskformer}
B.~Cheng, A.~G. Schwing, A.~Kirillov, Per-pixel classification is not all you need for semantic segmentation, in: Proc. Adv. Neural Inform. Process. Syst., Vol.~34, 2021, pp. 17864--17875.

\bibitem{MCN}
G.~Luo, et~al., Multi-task collaborative network for joint referring expression comprehension and segmentation, in: Proc. IEEE Conf. Comput. Vis. Pattern Recognit. (CVPR), 2020, pp. 10031--10040.

\bibitem{jain2023oneformer}
J.~Jain, J.~Li, M.~T. Chiu, A.~Hassani, N.~Orlov, H.~Shi, {OneFormer:} {One} transformer to rule universal image segmentation, in: Proc. IEEE Conf. Comput. Vis. Pattern Recognit. (CVPR), 2023, pp. 2989--2998.

\bibitem{MaskDINO}
F.~Li, et~al., Mask {DINO:} {Towards} a unified transformer-based framework for object detection and segmentation, in: Proc. IEEE Conf. Comput. Vis. Pattern Recognit. (CVPR), 2023, pp. 3041--3050.

\bibitem{wang2023hipie}
X.~Wang, S.~Li, K.~Kallidromitis, Y.~Kato, K.~Kozuka, T.~Darrell, Hierarchical open-vocabulary universal image segmentation, in: Proc. Adv. Neural Inform. Process. Syst., 2023.

\bibitem{UNINEXT}
B.~Yan, et~al., Universal instance perception as object discovery and retrieval, in: Proc. IEEE Conf. Comput. Vis. Pattern Recognit. (CVPR), 2023, pp. 15325--15336.

\bibitem{wu2023glee}
J.~Wu, Y.~Jiang, Q.~Liu, Z.~Yuan, X.~Bai, S.~Bai, General object foundation model for images and videos at scale, arXiv preprint arXiv:2312.09158 (2023).

\bibitem{liu2023universal}
Y.~Liu, C.~Zhang, Y.~Wang, J.~Wang, Y.~Yang, Y.~Tang, Universal segmentation at arbitrary granularity with language instruction, arXiv preprint arXiv:2312.01623 (2023).

\bibitem{sam}
A.~Kirillov, et~al., Segment anything, in: Proc. IEEE Int. Conf. Comput. Vis. (ICCV), 2023, pp. 4015--4026.

\bibitem{SEEM}
X.~Zou, et~al., Segment everything everywhere all at once, in: Proc. Adv. Neural Inform. Process. Syst., 2023.

\bibitem{ResNet}
K.~He, X.~Zhang, S.~Ren, J.~Sun, Deep residual learning for image recognition, in: Proc. IEEE Conf. Comput. Vis. Pattern Recognit. (CVPR), 2016, pp. 770--778.

\bibitem{ViT}
A.~Dosovitskiy, et~al., An image is worth 16x16 words: Transformers for image recognition at scale, in: Proc. Int. Conf. Learn. Represent. (ICLR), 2021.

\bibitem{BERT}
J.~Devlin, M.-W. Chang, K.~Lee, K.~Toutanova, {BERT}: {Pre-training} of deep bidirectional transformers for language understanding, in: Proc. Conf. North Amer. Chapter Assoc. Comput. Linguistics: Human Lang. Technol., Vol.~1, 2019, pp. 4171--4186.

\bibitem{bouchakour2018mfccs}
L.~Bouchakour, M.~Debyeche, {MFCCs} and gabor features for improving continuous arabic speech recognition in mobile communication modified, in: Proc. Int. Conf. Adv. Aspects Softw. Eng., Vol. 2326, 2018, pp. 115--121.

\bibitem{YOLOX}
Z.~Ge, S.~Liu, F.~Wang, Z.~Li, J.~Sun, {YOLOX}: Exceeding {YOLO} series in 2021, arXiv preprint arXiv:2107.08430 (2021).

\bibitem{CondInst}
Z.~Tian, C.~Shen, H.~Chen, Conditional convolutions for instance segmentation, in: Proc. Eur. Conf. Comput. Vis. (ECCV), Vol. 12346, 2020, pp. 282--298.

\bibitem{lin2017focal}
T.-Y. Lin, P.~Goyal, R.~Girshick, K.~He, P.~Doll{\'a}r, Focal loss for dense object detection, in: Proc. IEEE Int. Conf. Comput. Vis. (ICCV), 2017, pp. 2999--3007.

\bibitem{FasterRCNN}
S.~Ren, K.~He, R.~Girshick, J.~Sun, Faster {R-CNN:} {Towards} real-time object detection with region proposal networks, in: Proc. Adv. Neural Inform. Process. Syst., Vol.~28, 2015, pp. 91--99.

\bibitem{GIoULoss}
H.~Rezatofighi, N.~Tsoi, J.~Gwak, A.~Sadeghian, I.~Reid, S.~Savarese, Generalized intersection over union: A metric and a loss for bounding box regression, in: Proc. IEEE Conf. Comput. Vis. Pattern Recognit. (CVPR), 2019, pp. 658--666.

\bibitem{lin2023adaptiveclick}
J.~Lin, et~al., {AdaptiveClick:} {Clicks-aware} transformer with adaptive focal loss for interactive image segmentation, arXiv preprint arXiv:2305.04276 (2023).

\bibitem{DiceLoss}
F.~Milletari, N.~Navab, S.-A. Ahmadi, {V-Net:} {Fully} convolutional neural networks for volumetric medical image segmentation, in: Int. Conf. 3D Vis. (3DV), 2016, pp. 565--571.

\bibitem{han2023html}
M.~Han, Y.~Wang, Z.~Li, L.~Yao, X.~Chang, Y.~Qiao, {HTML:} {Hybrid} temporal-scale multimodal learning framework for referring video object segmentation, in: Proc. IEEE Int. Conf. Comput. Vis. (ICCV), 2023, pp. 13414--13423.

\bibitem{AdamW}
I.~Loshchilov, F.~Hutter, Decoupled weight decay regularization, in: Proc. Int. Conf. Learn. Represent. (ICLR), 2019.

\bibitem{tang2023temporal}
J.~Tang, G.~Zheng, S.~Yang, Temporal collection and distribution for referring video object segmentation, in: Proc. IEEE Int. Conf. Comput. Vis. (ICCV), 2023, pp. 15466--15476.

\bibitem{li2023robust}
X.~Li, J.~Wang, X.~Xu, X.~Li, B.~Raj, Y.~Lu, Robust referring video object segmentation with cyclic structural consensus, in: Proc. IEEE Int. Conf. Comput. Vis. (ICCV), 2023, pp. 22236--22245.

\bibitem{liu2022video}
Z.~Liu, J.~Ning, Y.~Cao, Y.~Wei, Z.~Zhang, S.~Lin, H.~Hu, Video swin transformer, in: Proc. IEEE Conf. Comput. Vis. Pattern Recognit. (CVPR), 2022, pp. 3202--3211.

\bibitem{roberta}
Y.~Liu, et~al., {RoBERTa:} {A} robustly optimized {BERT} pretraining approach, arXiv preprint arXiv:1907.11692 (2019).

\end{thebibliography}
%填写.bib文件的文件名
%\end{thebibliography}

%\clearpage
\end{document}